\documentclass{article}

\usepackage{microtype}
\usepackage{graphicx}
\usepackage{subfigure}
\usepackage{booktabs} % for professional tables
\usepackage{wrapfig}
\usepackage{xcolor}
\usepackage{pifont}
\usepackage{multirow}
\usepackage{hyperref}

\usepackage{amsmath}
\usepackage{amsthm}
\usepackage{amssymb}
\usepackage{enumitem}
\usepackage[accepted]{icml2021}

\usepackage{algorithmic}
\renewcommand{\algorithmicrequire}{\textbf{Input:}}
\renewcommand{\algorithmicensure}{\textbf{Output:}}
\definecolor{Tianlong_color}{rgb}{0.858, 0.188, 0.478}

\icmltitlerunning{Efficient Lottery Ticket Finding: Less Data is More}

\begin{document}

\twocolumn[
\icmltitle{Efficient Lottery Ticket Finding: Less Data is More}



\icmlsetsymbol{equal}{*}

\begin{icmlauthorlist}
\icmlauthor{Zhenyu Zhang}{equal,ustc}
\icmlauthor{Xuxi Chen}{equal,ut}
\icmlauthor{Tianlong Chen}{equal,ut}
\icmlauthor{Zhangyang Wang}{ut}
\end{icmlauthorlist}

\icmlaffiliation{ustc}{University of Science and Technology of China}
\icmlaffiliation{ut}{University of Texas at Austin}
\icmlcorrespondingauthor{Zhangyang Wang}{atlaswang@utexas.edu}

\icmlkeywords{Machine Learning, ICML}

\vskip 0.3in
]


\newcommand{\xuxi}[1]{\textcolor{red}{#1}}
\printAffiliationsAndNotice{\icmlEqualContribution} % otherwise use the standard text.

\begin{abstract}
The lottery ticket hypothesis (LTH) \cite{frankle2018lottery} reveals the existence of winning tickets (sparse but critical subnetworks) for dense networks, that can be trained in isolation from random initialization to match the latter's accuracies. However, finding winning tickets requires burdensome computations in the train-prune-retrain process, especially on large-scale datasets (e.g., ImageNet), restricting their practical benefits. This paper explores a new perspective on finding lottery tickets more efficiently, by doing so only with a specially selected subset of data, called \textbf{Pr}uning-\textbf{A}ware \textbf{C}ritical set (PrAC set), rather than using the full training set. The concept of PrAC set was inspired by the recent observation, that deep networks have samples that are either \textit{hard to memorize} during training, or \textit{easy to forget} during pruning. A PrAC set is thus hypothesized to capture those most challenging and informative examples for the dense model. We observe that a high-quality winning ticket can be found with training and pruning the dense network on the very compact PrAC set, which can substantially save training iterations for the ticket finding process. Extensive experiments validate our proposal across diverse datasets and network architectures. Specifically, on CIFAR-10, CIFAR-100, and Tiny ImageNet, we locate effective PrAC sets at $35.32\% \sim 78.19\%$ of their training set sizes. On top of them, we can obtain the same competitive winning tickets for the corresponding dense networks, yet saving up to $82.85\% \sim 92.77\%$, $63.54\% \sim 74.92\%$, and $76.14\% \sim 86.56\%$ training iterations, respectively. Crucially, we show that a PrAC set found is \textbf{reusable} across different network architectures, which can amortize the extra cost of finding PrAC sets, yielding a practical regime for efficient lottery ticket finding.
\end{abstract}

\section{Introduction}
\vspace{-0.3em}
Deep neural networks (DNNs) have revolutionized the performance bar of various tasks, yet suffer from substantial over-parameterization~\citep{voulodimos2018deep}. Parameter-counts are frequently measured in billions rather than millions, with the time and financial outlay necessary to train these models growing in concert. Once trained, they can be pruned of excessive capacity~\citep{han2015deep,tang2020scop}. However, conventional approaches first train dense DNNs, and then prune the trained them to high levels of sparsity. Those methods significantly reduce the inference complexity yet cost even greater computational resources and memory footprints at training. 

% Deep learning methods have outperformed previous state-of-the-art machine learning techniques in several fields over the past few years~\citep{voulodimos2018deep}. However there are bottlenecks that will obstruct the widely use of deep learning models. One of the obstacles is that deep neural networks are computationally intensive and require large memory, making them hard to deploy on devices with limited hardware resources~\citep{han2015deep}. %For instance, one single forward of ResNet-50~\citep{he2016deep}, a widely used deep learning model in computer vision, will require 4 GFlops. It was also estimated that NLP model training and development would make-up a substantial portion of the greenhouse gas emissions~\citep{strubell2019energy}. 
% Therefore, there is an increasing need of efficient training methods that can save model performance to the greatest extent. 

\begin{figure}
\vspace{-1em}
\centering
\includegraphics[width=1\linewidth]{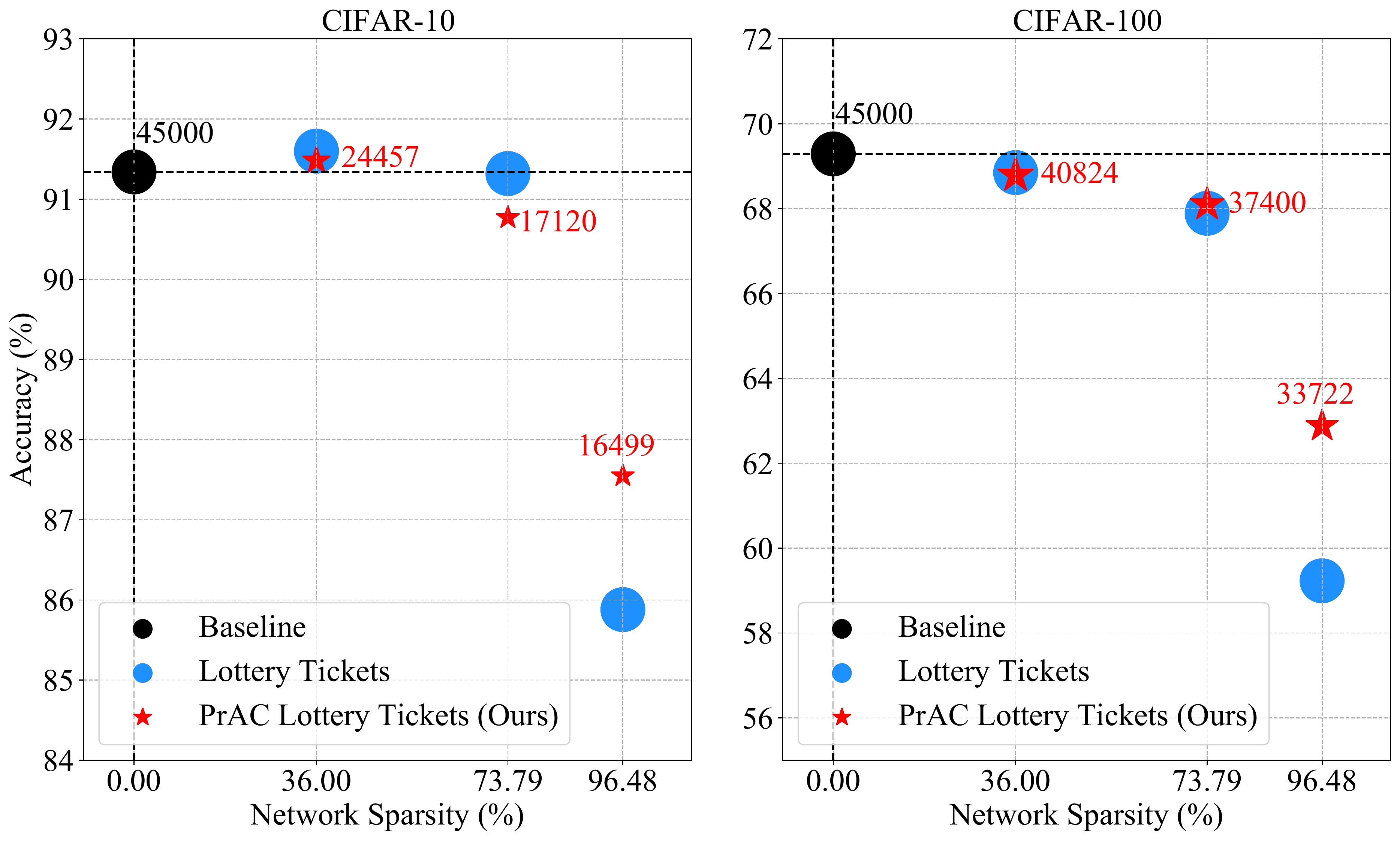}
\vspace{-6mm}
\caption{Test accuracy of found subnetworks from ResNets at different sparsity levels on CIFAR-10 and CIFAR-100. Black dots ($\bullet$) represent the performance of unpruned baselines; blue dots ($\textcolor{blue}{\bullet}$) indicate the performance of vanilla lottery tickets found with full data \cite{frankle2018lottery}, and red star (\textcolor{red}{\ding{72}}) are established by our PrAC lottery tickets. \textcolor{red}{Red} numbers are the number of samples in the PrAC set. The lottery tickets found on the PrAC sets could perform on par with the vanilla ones at moderate sparsity levels, and even outperform the latter at the highest sparsity of 96.48\%.}
\vspace{-1.5em}
\label{fig:teaser}
\end{figure}

An emerging subfield has explored the prospect of directly training smaller, sparse subnetworks in place of the full models without sacrificing performance. The key idea is to reuse the sparsity pattern found through pruning and train a sparse network from scratch. The seminal work~\citep{frankle2018lottery} hypothesized that standard DNNs contain sparse matching subnetworks, often called $\textit{winning ticket}$, capable of training in isolation to full accuracy. In other words, we could have trained smaller networks from the start if only we had known which subnetworks to choose. In larger-scale real-world settings, current methods often empirically choose winning tickets by \textit{Iterative Magnitude Pruning} (\textbf{IMP}), sometimes at an early training point called ``rewinding" \citep{frankle2019stabilizing,frankle2020linear}. Other works also showed sparsity might emerge at the initialization \cite{lee2018snip,Wang2020Picking}, or at the early training stage \cite{You:2019tz}. However, it was observed in \cite{frankle2020pruning} that IMP still outperforms those carefully designed alternatives by clear margins, and remain as the most effective lottery ticket finding approach. However, the cumbersome train-prune-train cycle required by IMP makes it extremely expensive to find lottery tickets from large models and datasets, and also questioning the practical efficiency benefits of finding lottery tickets.

In parallel to seeking \textit{model sparsity} during training, another complementary and promising line of ideas exploits \textit{data sparsity}, \textit{i.e}, reducing training costs by the informed selection of training samples~\citep{tsang2005core,har2007smaller}. Such techniques often select a small but critical \textit{core set} from a large dataset, by which way a significant fraction of examples can be omitted from training while still maintaining the trained models' generalization  \cite{zhao2015stochastic,katharopoulos2018not,toneva2018an,mirzasoleiman2020coresets}. Also related to the core set approach is the dataset distillation \citep{wang2018dataset} that aims to summarize training images into a handful of synthetic images, ensuring that DNNs trained on the latter generalize almost as well as trained on the former. 

\vspace{-0.5em}
\subsection{Research Questions \& Our Contributions}
\vspace{-0.2em}
However, the questions below are not yet clear:
\begin{center}
\vspace{-0.5em}
\textit{(Q$_1$) How will the ``model sparsity" (e.g. LTH) and ``data sparsity" (e.g., core set) interplay? Can one help the other? Can they possibly be jointly utilized to push training efficiency to the next level?}
\vspace{-0.5em}
\end{center}
To answer the above question (Q$_1$), we first formulate and address a prerequisite question (Q$_0$): 
\begin{center}
\vspace{-0.5em}
\textit{(Q$_0$) What samples are considered as ``core" for finding a lottery ticket (trainable sparse DNN)?}
\vspace{-0.5em}
\end{center}
A typical ``coreset" \cite{mirzasoleiman2020coresets} aims to guarantee
that models fitting the coreset also provide a good fit for the original data, and finding it is treated as an approximation problem such as sampling or clustering. To find a sparse subnetwork that can \textit{match} the performance of the full model, the challenge level is escalated higher since sparse DNNs are way tougher to train~\citep{evci2019difficulty}, and the core samples have also to identify the trainable sparse connectivity patterns. In other words, the new core set needs to encode not only the full dataset's knowledge, but also the trainability. 

In this paper, we first attempt to address (Q$_0$) by investigating a new concept called \textbf{Pr}uning-\textbf{A}ware \textbf{C}ritical set (\textbf{PrAC set}), that targets to characterize important samples for finding lottery tickets that are both same \textit{generalizable} and \textit{trainable}. Considering that the lottery ticket iterates between two steps: \textit{(re-)training}, and \textit{pruning}. Conceptually, we hope a PrAC set to capture two types of samples:
\begin{itemize} 
\vspace{-1em}
    \item Samples that are \textit{hard to memorize}, during (re-)training of the (original or pruned) DNN. Recent observations by \cite{toneva2018an,yao2020searching,xia2021robust,han2020sigua} reveal that certain examples are memorized easily during training, but some others are repeatedly forgotten. Such (un)forgettable examples generalize across different architectures in the same dataset. The forgetting dynamics also suggest one can train a DNN on a dataset with a large fraction of the least forgotten examples removed.\vspace{-0.5em} 
    \item Samples that are \textit{easy to forget}, during pruning the dense DNN into a sparse DNN. Pruning steps are essential to the final (trainable) sparsity, yet hampering both memorization and generalization. Moreover, it has been observed by \cite{hooker2020compressed} that pruning disproportionately impacts the model performance on a narrow subset of the dataset, e.g., the atypical, semantically ambiguous or underrepresented images.\vspace{-1em}

\end{itemize}
During lottery ticket finding, by calculating the forgotten dynamics for each sample within training and the prediction differences after each pruning, we can effectively collect those most informative samples and build a PrAC set. In fact, our approach is a \textit{co-design between data and model sparsity}, which we feel essential due to the hardness of (Q$_0$). 

% How overlapping are the two sets?
% Are they complementary? 

Equipped with PrAC sets, we then examine (Q$_1$) and present a comprehensive set of experiments, integrating the PrAC set with an efficient lottery ticket finding and training framework. In general, we find PrAC sets to help find comparable winning tickets with much higher training efficiency, compared to the vanilla IMP scheme using the full set, with little performance drop (sometimes even with performance gains)\footnote{Our implementations are available at: \url{https://github.com/VITA-Group/PrAC-LTH}}. 
%We validate the importance of the PAC set by comparing with multiple sample selection methods. We also compare our framework with state-of-the-art efficient pruning approaches~\citep{lee2018snip,Wang2020Picking,tanaka2020pruning} as well as the random pruning baseline~\citep{chen2020lottery}. 
We summarize our main findings as follows:\vspace{-1em} 
\begin{itemize}[leftmargin=*]
    \item We identify winning tickets and PrAC sets broadly across different datasets (CIFAR-10, CIFAR-100, Tiny ImageNet) and architectures (ResNet-20, ResNet-56, and VGG-16). High-quality winning tickets can be found on the PrAC sets while saving training time and costs. Specifically, we save $82.85\% \sim 92.77\%$ on CIFAR-10, $63.54\%\sim74.92\%$ on CIFAR-100, and $76.14\% \sim 86.56\%$ on Tiny ImageNet in training iterations, while maintaining or even boosting their achievable accuracies.\vspace{-0.5em} 
    \item PrAC sets show great transferability across architectures on the same dataset, which can amortize the cost of finding PrAC sets in practice. Taking ResNet-20 as the source architecture, the PrAC set found in CIFAR-10 and CIFAR-100 can locate winning tickets in ResNet-56 and VGG-19 with almost no performance degradation. We further visualize the PrAC set samples, conclude their patterns, and compare them with multiple sample selection methods.\vspace{-0.5em}  
    \item On CIFAR-10, the PrAC winning ticket (79.03\%) are sparser than tickets from random pruning (48.80\%). Our ticket finding also outperforms other efficient network pruning methods. For example, at 93.13\% sparsity, our PrAC lottery tickets can outperform SynFlow~\citep{tanaka2020pruning} by 1.51\%, SNIP~\citep{lee2018snip} by 5.47\%, and GraSP~\citep{Wang2020Picking} by 18.73\%.\vspace{-0.5em}  
\end{itemize}

\vspace{-1.2em}  
\section{Related Work}
\vspace{-0.2em}  
\paragraph{Lottery Ticket Hypothesis (LTH).} LTH \citep{frankle2018lottery} has drawn lots of attention. Later on, \citep{frankle2019linear,renda2020comparing} scaled up LTH to larger models by early weight rewinding that relaxes the use of original random initialization. Another intriguing property of lottery tickets, the transferability, has also been thoroughly examined \citep{mehta2019sparse,morcos2019one,desai2019evaluating,chen2020lottery,chen2020lottery2}. \citet{zhou2019deconstructing} investigated different components in LTH and observed super-masks in winning tickets. LTH has also been extended to various applications~\cite{gale2019state,chen2020lottery,yu2019playing,chen2021gans, kalibhat2020winning, chen2021ultra,ma2021good,gan2021playing,chen2021unified} beyond image classification.

%The original version of lottery ticket hypothesis was unable to scale up the large models, but such disadvantage has been overcame by early weight rewinding \citep{frankle2019linear,renda2020comparing} that relaxed the requirement of using the same initialization. Other intriguing properties of lottery tickets like transferability, also been throughout exploited \citep{mehta2019sparse,morcos2019one,desai2019evaluating,chen2020lottery,chen2020lottery2}. \citet{zhou2019deconstructing} investigated different components in LTH and observed super-masks in winning tickets. Meanwhile, LTH has been broadly extended to various fields including image classification \cite{liu2018rethinking, evci2019difficulty, savarese2020winning,You:2019tz}, image generation \citep{chen2021gans, kalibhat2020winning}, natural language processing \cite{gale2019state,chen2020lottery}, and reinforcement learning \cite{yu2019playing}. 

Unstructured IMP \citep{han2015deep,frankle2018lottery} serves as an effective method to find these winning tickets, and Dynamic Sparse Training~\citep{mostafa2019parameter,mocanu2018scalable,evci2020rigging} is also capable of identifying subnetworks with promising performance. However their computational expensiveness motivates many efficient alternatives that hope to locate sparse trainable subnetworks at random initialization or early training stage, with less or no training \citep{lee2018snip,You:2019tz,Wang2020Picking,tanaka2020pruning,frankle2020pruning}. Unfortunately, those sparse subnetworks found at beginning usually have clearly \textit{inferior performance} to the IMP-found winning tickets, leaving IMP still the mainstream LTH scheme. This paper explores a complementary new perspective on finding lottery tickets more efficiently by co-designing a specially crafted subset. Our method secures winning tickets of \textit{fully comparable performance} to the full IMP scheme, and it can also be straightforwardly combined with those efficient pruning methods if needed.

%However, to our best knowledge, efficient ticket-finding has been less explored. \citep{lee2018snip,Wang2020Picking,tanaka2020pruning} prune networks at random initialization, which pursue efficient training at the cost of scarified performance. \citet{You:2019tz} is another pioneer to introduce early stopping and low-precision training towards improved training efficiency. 

\vspace{-1em}  
\paragraph{Active Learning and Core-Set Approaches.} Another closely related literature is the problem of active learning~\citep{settles2009active,settles2012active} and core-set selection~\citep{tsang2005core, har2007smaller, bachem2017practical, sener2017active}. Specifically, \citet{zhao2015stochastic, katharopoulos2018not,toneva2018an,wang2018dataset,mirzasoleiman2020coresets,hooker2020compressed,hooker2020characterising} select core-sets by the importance sampling. \citet{zhao2015stochastic, katharopoulos2018not} sort the samples according to the magnitude of its loss gradient with respect to parameters of the network. \citet{toneva2018an} samples the examples based on the forgetting dynamics during the course of learning. \citet{mirzasoleiman2020coresets} constructs core-set that provides an approximately low-rank Jacobian matrix. \citet{wang2018dataset} generates synthetic examples to distill the knowledge from the entire dataset, and \citet{hooker2020compressed,hooker2020characterising} find pruning can cause disproportionately high errors on a small subset. We draw inspirations from several of those ideas, and extend the idea of core-set to be co-optimized with LTH.

%It can be roughly divided into two groups: pool-based~\citep{yang2003automatically,tur2005combining,hauptmann2006extreme} and stream-based active learning~\citep{thompson1999active,moskovitch2007improving}. As for core-set selection~\citep{tsang2005core, har2007smaller, bachem2017practical, sener2017active}, it summarizes the whole dataset into a small subset such that the model trained on the subset can achieve similar performance as trained on the entire dataset. 

\vspace{-1em}  
\section{Methodology}
\vspace{-0.5em}  
In this section, we present our framework to co-design model and data sparsity, which works in an iterative fashion of two alternative steps: i) constructing the Pruning-Aware Critical (PrAC) set with pruned models, which selects the most challenging and informative examples; ii) utilizing PrAC sets to identify critical subnetworks, (\textit{i.e.}, lottery tickets), which takes much less training iterations. In this way, the burdensome computations of the train-prune-retrain process in tickets finding, can be substantially reduced. The overall pipeline is summarized in Algorithm~\ref{alg2}.

\vspace{-0.8em}  
\begin{algorithm}
\caption{Data and Model Sparsity Co-Design} 
\label{alg2}
\begin{algorithmic}[1]
\REQUIRE Full training data $\mathcal{D}_0$, a threshold for the number of forgets $\mathcal{E}_{\mathrm{F}}$, a network $f(\boldsymbol{\theta}_0,)$ with initialization weights $\boldsymbol{\theta}_0$, pruning ratios $\rho$, and the desired sparsity level $s$.
\ENSURE Sparse mask $\boldsymbol{m}$ ($\|\boldsymbol{m}\|_0\ll\|\boldsymbol{\theta}_0\|$), pruning-aware critical (PrAC) set $\mathcal{P}$ ($|\mathcal{P}|\ll|\mathcal{D}_0|$)
\STATE Set $\boldsymbol{m} = \boldsymbol{1}\in\mathbb{R}^{\|\boldsymbol{\theta}_0\|_0}$, and $\mathcal{D}=\mathcal{D}_0$
\WHILE {$(1-\frac{\|\boldsymbol{m}\|_0}{\|\boldsymbol{\theta}_0\|_0} \le s )$}
\STATE \textcolor{gray}{\# Data slimming to construct PrAC sets}
\STATE Set $\mathcal{P}=\varnothing$
\STATE Train $f(\boldsymbol{m}\odot\boldsymbol{\theta}_0,\cdot)$ on $\mathcal{D}$ for $\mathrm{T}$ epochs and update the forgetting statistics for all training samples in $\mathcal{D}$
\STATE Select samples from $\mathcal{D}$ with forgetting statistics greater than $\mathcal{E}_\mathrm{F}$, and add them into $\mathcal{P}$
\STATE \textcolor{gray}{\# Model slimming to locate critical subnetworks}
\STATE Prune $\rho=20\%$ remaining weights of subnetworks $f(\boldsymbol{m}\odot\boldsymbol{\theta}_{\mathrm{T}},\cdot)$, and update $\boldsymbol{m}$ accordingly
\STATE \textcolor{gray}{\# Data slimming to construct PrAC sets}
\STATE Select samples from $\mathcal{D}_0$ that full model and subnetworks \textbf{disagree with}, and add them into into $P$
\STATE Set $\mathcal{D}=\mathcal{P}$ 
\ENDWHILE 
\end{algorithmic}
\end{algorithm}
\vspace{-1em}

\vspace{-0.5em}  
\subsection{Identifying the Pruning-Aware Critical (PrAC) Set}
\vspace{-0.5em}  
\label{sec:pac}
This section shows the details about how to shrink the training set to proposed Pruning-Aware Critical set, which illustrates the process in lines 3-6, 10 of Algorithm~\ref{alg2}.

\vspace{-0.5em}  
\paragraph{Rationale I: Critical Examples for Training.} In the network training, each batch of data has its own and likely different statistics. Therefore, they can be regarded as different ``tasks''. Catastrophic forgetting happens~\citep{toneva2018an} during the training process so that %there will always be samples that models are forgetting, have forgotten, or never forget. %
certain examples are memorized easily during training while some others are repeatedly forgotten. Different behaviors on samples reveal the difficulty of them, providing a natural way to select critical examples, \textit{i.e.}, the degree of difficult-to-forget of each training sample. As pointed out by \citep{toneva2018an}, training models on a dataset with a large fraction of the least forgotten examples removed can yield extremely competitive performance as training on the full data.  

\vspace{-0.9em}  
\paragraph{Approach I: Calculating the Forgetting Statistics.} To measure how easy for a model to forget a sample, we use \textit{forgetting statistics}~\citep{toneva2018an} as the metric. Specifically, the forgetting statistics for a sample is the number of transition from a correctly to incorrectly classified sample. We sort the number of forgetting statistics of all training data, and select those have statistics greater than a pre-defined threshold into the PrAC set, in lines 3-6 of Alg.~\ref{alg2}. 

\vspace{-0.9em}  
\paragraph{Rationale II: Critical Examples for Pruning.} Although the performance of located sparse lottery tickets can match the performance of the full model, the increased number of zero weights might have hampered the memorization and generalization ability of models. Such conjecture has been supported by recent observation~\citep{hooker2020compressed}, which demonstrates there exists pruning-aware examples that have different prediction between the full and pruned model. These examples are semantically ambiguous and hard for the pruned model to memorize.%, which can reflect the mysterious capacity loss after pruning. 
As a consequence, we merge these easy-to-forget samples into the PrAC set we construct to remedy such capacity loss. 

\vspace{-0.9em}  
\paragraph{Approach II: Utilizing the Disagreement between Full and Pruned Models.} For each sample in the training set, we calculate the predicted class of $\boldsymbol{x}$ by full dense models and pruned subnetworks, \textit{i.e.}, $f(\boldsymbol{\theta}, \boldsymbol{x})$ and $f(\boldsymbol{m} \odot \boldsymbol{\theta}, \boldsymbol{x})$, where $f(\boldsymbol{\theta}, \cdot)$ is a model with parameters $\boldsymbol{\theta}$, and $\boldsymbol{m}$ is a sparse mask. If two predictions are different, then we include this sample to the PrAC set (line 10 of Algorithm~\ref{alg2}). 

\vspace{-0.5em}  
\subsection{Efficient Lottery Tickets Finding}
%In this section, we detail the pipeline to find lottery tickets efficiently, in terms of training iterations. 
\vspace{-0.2em}  
\paragraph{Matching Subnetworks and Lottery Ticket.} A subnetwork within a dense network $f(\boldsymbol{\theta}, \cdot)$ is defined as $f(\boldsymbol{m} \odot \boldsymbol{\theta}, \cdot)$, where $\boldsymbol{m} \in \{0,1\}^{\|\boldsymbol{\theta}\|_0}$ is a binary mask indicating the sparsity levels, and $\odot$ is the element-wise product. Let $\boldsymbol{\theta}_0$ be the initial weights, and $\boldsymbol{\theta}_i$ be the weights after $i$ training steps. Following \citet{frankle2018lottery}, we define the \textit{matching network} as a subnetwork $f(\cdot,\boldsymbol{m}\odot\boldsymbol{\theta})$, with $\boldsymbol{\theta}_t$ being the initialization of $\boldsymbol{\theta}$, that can reach the comparable performance to the full network within a similar training iterations;
a \textit{winning ticket} is defined as a matching subnetwork %$f(\cdot; \boldsymbol{m}\odot\boldsymbol{\theta})$ where %$\boldsymbol{\theta} = \boldsymbol{\theta_0}$. 
where $\boldsymbol{\theta_0}$ as the initial weights.

\vspace{-0.5em}  
\paragraph{Identifying Subnetworks.} To identify subnetworks, we adopt an iterative magnitude pruning method \cite{han2015deep}. We follow a conventional iterative train-prune-retrain process in \citet{frankle2018lottery}, yet with our PrAC set: We train the model $f(\boldsymbol{m} \odot \boldsymbol{\theta}, \cdot)$ on our PrAC set, prune a certain percent of the weights, reset and retrain the model, and repeat the process until we meet the sparsity requirement. 

%We study two different retraining protocols: re-train with full data and re-train with the PAC set. Re-training the model on our PAC set can save training iterations since we 

%\textcolor{blue}{Although re-training on PAC set yields slightly inferior results, such re-training protocol can gain great training efficiency.} 

%\paragraph{Early Bird Tickets} \cite{You:2019tz} claim that winning tickets emerge in the early period of training process. We implement the proposed early bird tickets in our methods. We perform unstructured pruning on current network every fixed number of iterations and calculate the hamming distance between the last and current mask. Once the mask distance become small enough, we find the early bird ticket.
\begin{figure}[t]
    \centering
    \includegraphics[width=1\linewidth]{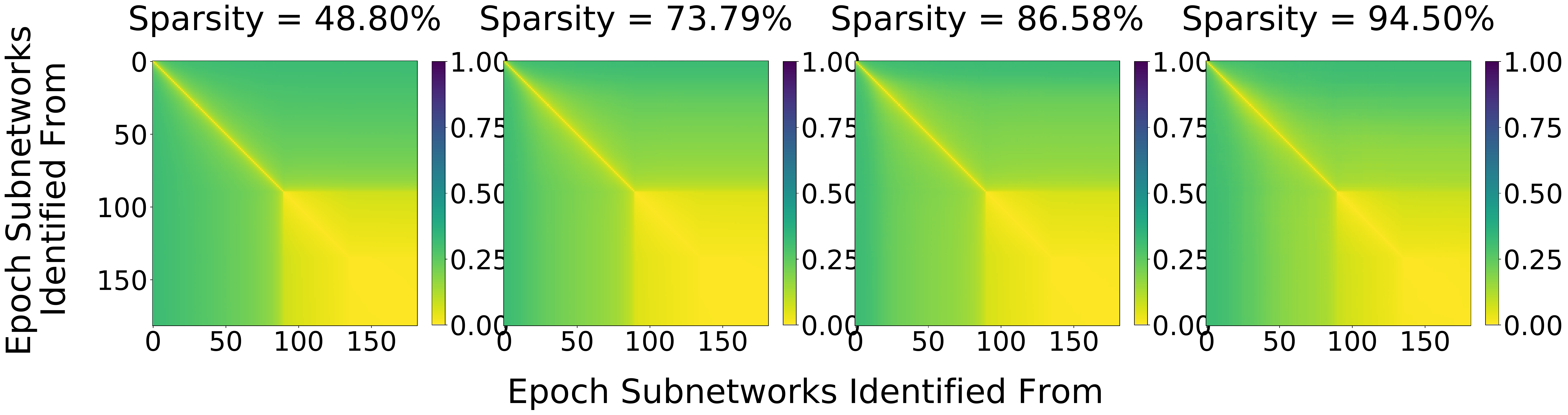}
    \vspace{-8mm}
    \caption{Results of the pairwise hamming distance between identified subnetworks on CIFAR-10 with ResNet-20}
    \label{fig:eb_mask_heatmap}
    \vspace{-4mm}
\end{figure}

\vspace{-0.5em}  
\paragraph{Turning PrAC set into actual training efficiency.} Using PrAC sets for training can save training cost, firstly because less training data directly lead to fewer training iterations per training epoch. However, \textbf{the gains are way beyond linear} - since less training data could also imply easier fitting and faster convergence, e.g., less number of epochs. To fully leverage the potential of PrAC sets for efficient ticket finding, we introduce two training strategies for PrAC: 
\begin{itemize} [leftmargin=*]
\vspace{-1em}  
    \item [i)] \textbf{\textit{Dynamic training iterations}}. After constructing the PrAC set, we will tune the training iterations according to the size of the PrAC set. We linearly scale down the number of iterations using the following formula to decide a new number of training iterations: $ N = \frac{|\mathcal{P}|}{|\mathcal{D}_0|}N_0$, where $\mathcal{D}_0$ and $\mathcal{P}$ are the full training set and the PrAC set respectively, and $N_0$ is the original training iterations. 
    
    In practice, we also tune the learning rate scheduler using the above adjustment formula to re-calculate the decay schedule for learning rates. By scaling down the required training iterations, we can gain training efficiency in a simple but meaningful way.\vspace{-0.5em}  
    \item [ii)] \textbf{\textit{Early stopping}}. We build an early stopping mechanism upon the dynamic training iterations technique by introducing the Early Bird Ticket~\citep{You:2019tz}. It was originally designed for one-shot pruning; however, we reformulate and extend it to our iterative pruning context. As shown by \citet{You:2019tz}, winning tickets will emerge at the early period of the training process, which provides empirical support for using the early stopping technique. In our work, we calculate sparsity masks for the model after every epoch of training and monitor the distance between masks as a criterion for early stopping. 
    
    The distance metric for matrices we use is the Hamming distance, \textit{i.e.}, the number of different elements in two masks. Once the distance becomes smaller than a threshold, we interrupt the training, prune the network and update the sparsity mask, and use it for further retraining. The Hamming distances between masks at different sparsities on different architectures are shown in Figure~\ref{fig:eb_mask_heatmap}. The graph validates the convergence of Hamming distance between sparsity masks at about half of training.\vspace{-0.5em}    
\end{itemize}

\begin{figure*}[!htb]
    \centering
    \includegraphics[width=1\linewidth]{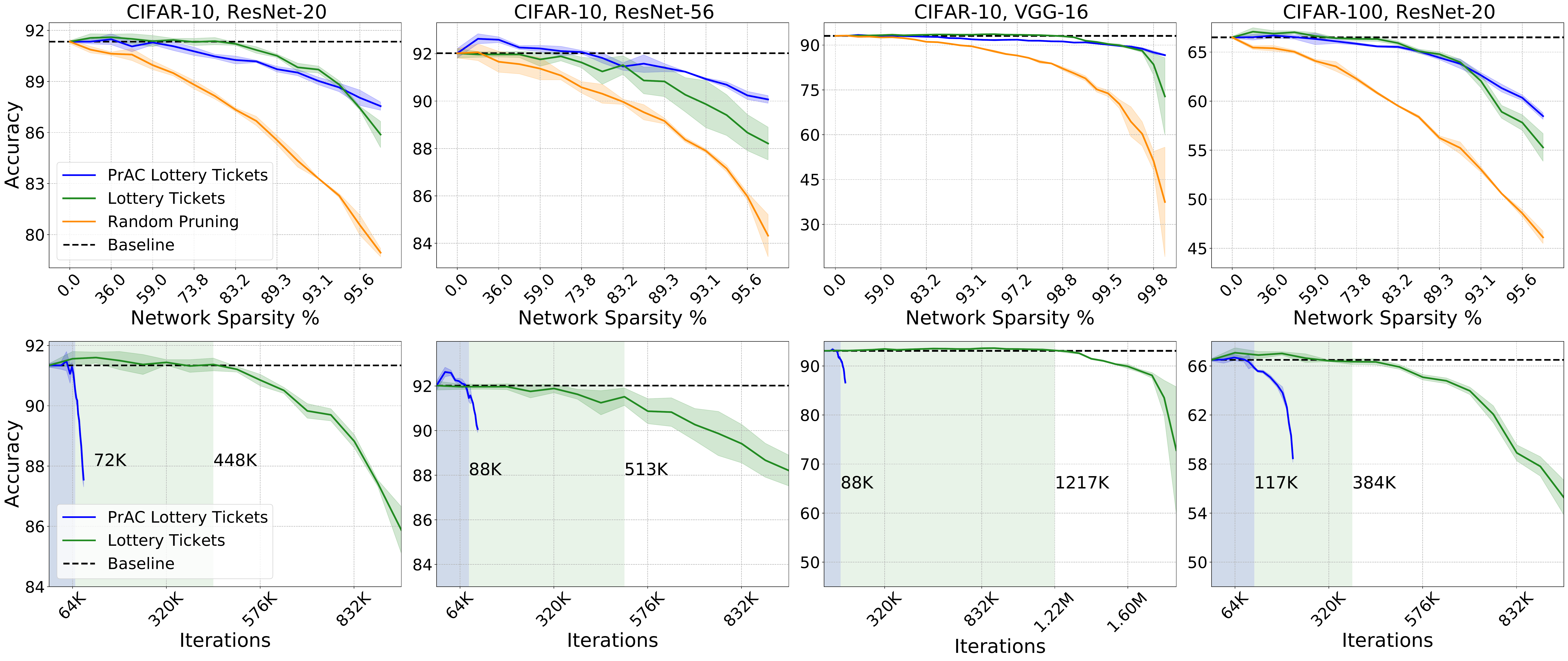}
    \includegraphics[width=1\linewidth]{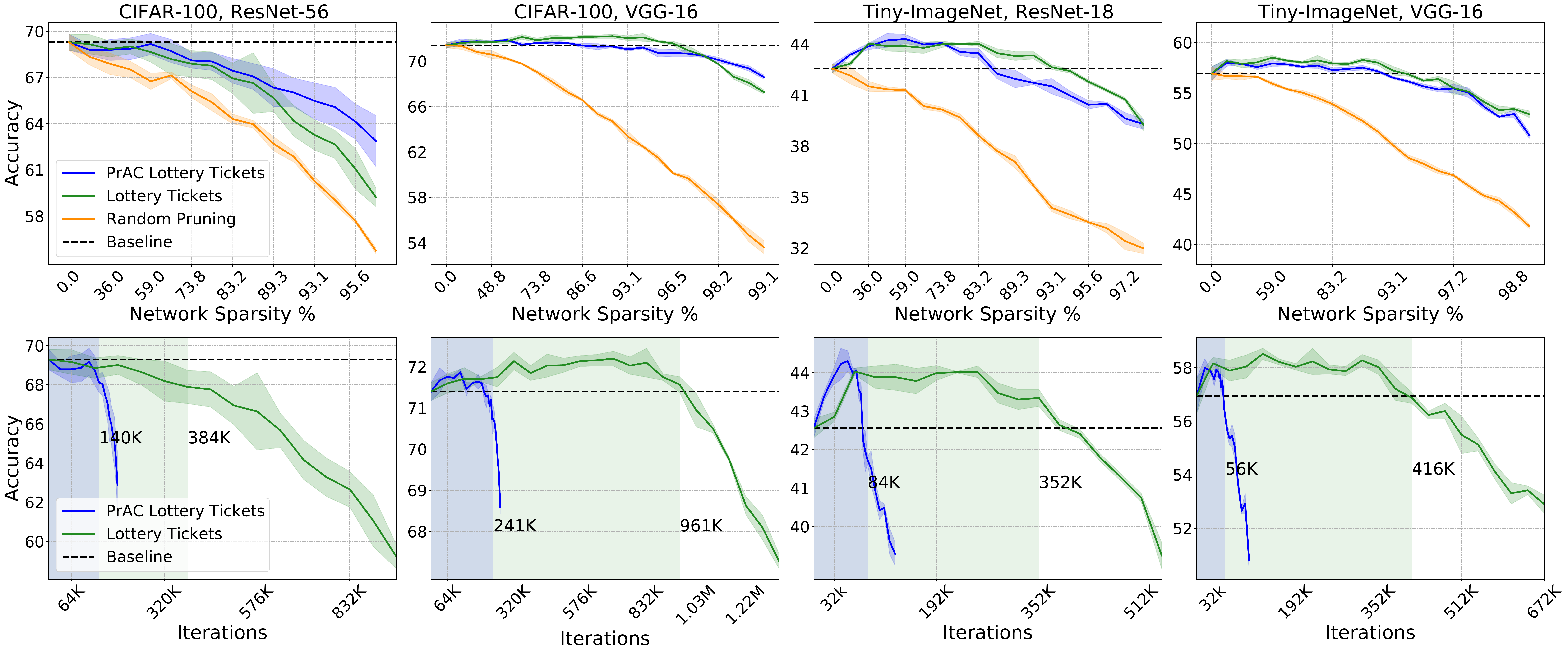}
    \vspace{-8mm}
    \caption{Testing accuracy of subnetworks at a range of sparsity levels from $0\%$ to $99.85\%$ (the first and third rows) and the training iterations for finding each subnetwork (the second and fourth rows) on CIFAR-10, CIFAR-100, and Tiny-ImageNet with ResNet-18, ResNet-20, ResNet-56, and VGG-16. \textcolor{blue}{Blue}, \textcolor{green}{Green}, \textcolor{orange}{Orange} and \textbf{Black} curves represent our PrAC lottery tickets, vanilla lottery tickets, random pruning, and dense network, respectively. The solid line and shading are the mean and standard deviation of testing accuracy. The numbers within figures are the iterations used to find subnetworks with the \textbf{same sparsity} and \textbf{comparable performance}, which indicate our achieved training resources saving. We consider PrAC lottery tickets to achieve a matched performance as vanilla lottery tickets when the performance of PrAC lottery tickets is within one standard deviation of the performance of vanilla lottery tickets.}
    \label{fig:main_result}
    \vspace{-4mm}
\end{figure*}

Integrating the above two techniques with the PrAC set, we build our data-model sparsity co-design framework to efficiently find matching subnetworks, termed as \textit{PrAC lottery ticket}, with much less training resources.

\vspace{-0.5em}  
\section{Experiments}
\vspace{-0.2em}  
\label{para:setup}

\begin{table}[!ht]
    \centering
    \vspace{-2mm}
    \caption{Implementation Details. For ResNet-20 and ResNet-56, we adopt three different training settings: standard, \textit{low} and \textit{warmup}~\cite{frankle2019linear}. The low variant means a lower learning rate, and the warmup variant adopts a warm-up method that linearly increases the learning rate from zero.}
    \label{tab:setting}
    \resizebox{0.48\textwidth}{!}{
    \begin{tabular}{cc|c|c|c|c}
        \toprule
         Network & Variant & Dataset & Batch Size & Learning Rate & Warmup  \\ \midrule
          & Standard & & & $0.1$ & $0$   \\
         ResNet-20 & Low & CIFAR10 \& CIFAR100 & $128$ & $0.01$ & $0$  \\
          & Warmup & & & $0.03$ & $15$ epochs  \\ \midrule
          & Standard & & & $0.1$ & $0$   \\
         ResNet-56 & Low & CIFAR10 \& CIFAR100 & $128$ & $0.01$ & $0$  \\
          & Warmup & & & $0.03$ & $15$ epochs  \\ \midrule
         \multirow{2}*{VGG-16}& - & CIFAR10 \& CIFAR100 & $128$ & $0.1$ & $0$ \\
		~ & - & Tiny-ImageNet & $512$ & $0.1$ & $0$ \\ \midrule
         ResNet-18& - & Tiny-ImageNet & $512$ & $0.1$ & $0$ \\ \bottomrule
    \end{tabular}}
    \vspace{-5mm}
\end{table}

\paragraph{General Setup.} 
We summarize the key setups and hyperparameters of our implementation in Table~\ref{tab:setting}, and refer readers to Appendix~\ref{appendix_setting} for more details. Our experiments use two popular architectures, ResNet~\citep{he2016deep} and VGG~\citep{simonyan2014very}, on three representative datasets, \textit{i.e.}, CIFAR-10~\citep{krizhevsky2009learning}, CIFAR-100~\citep{krizhevsky2009learning} and Tiny-ImageNet~\citep{wu2017tiny}. Specifically, we train networks for $182$ epochs with a multi-step learning rate schedule, which decays the learning rate to its one-tenth at epoch $91$ and $136$, respectively. We evaluate the quality of obtained subnetworks, \textit{i.e.}, lottery tickets, by testing accuracy after independently trained from the same random initialization or early rewound weights \cite{frankle2019linear}. All reported results are averaged over three independent runs.

\vspace{-0.5em}  
\subsection{Identifying Winning Tickets with PrAC Sets}
\vspace{-0.5em}  
We evaluate our data and model sparsity co-design framework across diverse datasets and architectures with a total of eight combinations, specifically, CIFAR-10 with \{ResNet-20, ResNet-56, VGG-16\}, CIFAR-100 with \{ResNet-20, ResNet-56, VGG-16\}, and Tiny-ImageNet with \{ResNet-18, VGG-16\}. We consider vanilla lottery tickets (LT) method~\citep{frankle2018lottery} and random pruning for comparisons. Figure~\ref{fig:main_result} collects the achieved performance of subnetworks with different sparsity and their training effort for identifying each subnetwork, in terms of training iterations. Several observations can be drawn as follows:
\begin{itemize} [leftmargin=*]
\vspace{-0.5em}  
    \item Our PrAC lottery tickets match the performance as vanilla lottery tickets in all combinations while notably less training costs, specifically achieving training iteration saving of $83.93\%$ and $69.53\%$ for ResNet-20, $82.85\%$ and $63.54\%$ for ResNet-56, $92.77\%$ and $74.92\%$ for VGG-16 on CIFAR-10 and CIFAR-100, respectively; $76.14\%$ for ResNet-18, and $86.56\%$ for VGG-16 on Tiny-ImageNet. As shown in Figure~\ref{fig:main_result}, we record the number of training iterations for the PrAC lottery tickets and the vanilla LT before reaching the highest sparsity that the former can match. And we color the area of the graph according to the number of training iterations for better demonstration.\vspace{-0.5em}    
    \item Somehow surprisingly, PrAC lottery tickets can even outperform the vanilla lottery tickets at some very high sparsity levels. This intriguing phenomenon implies that utilizing the data-level sparsity by PrAC sets, in addition to efficiency purpose, may even have additional regularization effects on improving the found model's generalization. We will leave further investigation for future work.
    %Given that our method use much less training data ($35.32\% \sim 78.19\%$, more details are listed in Table \ref{tab:pac_size}), a marginal performance gap of less than $1\%$ of testing accuracy does not outweigh our efficiency gain.
    % \vspace{-0.5em}   
    % \item \TL{PAC lottery tickets achieve almost matched performance (errobar are overlapped?)... not drop performance, how many iteration saving, what is the size of obtained PAC sets.} We observe that our proposal can find high-quality matching subnetworks as good as LT at non-trivial sparsity but save lots of training iterations ($92.77\%$ on CIFAR-10, $74.92\%$ on CIFAR-100 and XX\% on Tiny-ImageNet with VGG-16). Taking ResNet-20 on CIFAR-10 as an example, our method uses only \textbf{$\boldsymbol{16.07\%}$ training effort} and finds almost same good matching subnetworks. The testing accuracy of the identified subnetwork with $79.03\%$ sparisity by our approach and LT is $90.47\%$ and $91.37\%$, respectively. Given that our method use much less training data ($25.14\%$ data on CIFAR-10, $61.94\%$ data on CIFAR-100, and XX\% data on Tiny ImageNet with VGG-16), 
    %\item \TL{Analyses across datasets, different dataset have different observations}
    \item The numbers of examples in PrAC sets across different datasets are adaptively varying. On CIFAR-10, the percentage of the number of the PrAC sets ranges from $35.32\%$ to $37.07\%$, from $69.55\%$ to $78.19\%$ on CIFAR-100, and from $68.23\%$ to $75.10\%$ on Tiny ImageNet. The ratios of training iterations saved also vary between datasets. On CIFAR-10, we can save training iterations more than $80\%$ but no more than $75\%$ on CIFAR-100, which means that it requires more training effort to find PrAC lottery ticket on CIFAR-100 than CIFAR-10.\vspace{-0.5em}   
    \item Different architectures show the different percentage of training iteration saving and indicates the speed of matching subnetworks emerge. On VGG-16, our method can save the highest percentage of training iterations, indicating the highest speed to find lottery tickets. On ResNet-56 and ResNet-18, the speed to find lottery ticket is slower; On CIFAR-10 our framework can save $82.85\%$ of training iterations on ResNet-56 while $92.77\%$ on VGG-16; On Tiny ImageNet our framework can save $76.14\%$ on ResNet-18 while $86.56\%$ on VGG-16.\vspace{-0.5em}   
    %\item \TL{Analyses across architectures}
\end{itemize}

% Table~\ref{tab:main_result} shows that with only $10\%$ training effort, our method  We also show the performance of subnetworks with different sparsity in Figure~\ref{fig:main_curve_3}. Our method can reduce training effort but achieve similar performance as vanilla IMP.  

\begin{figure}[t]
% \vspace{-0.5em} 
    \centering
    \includegraphics[width=1\linewidth]{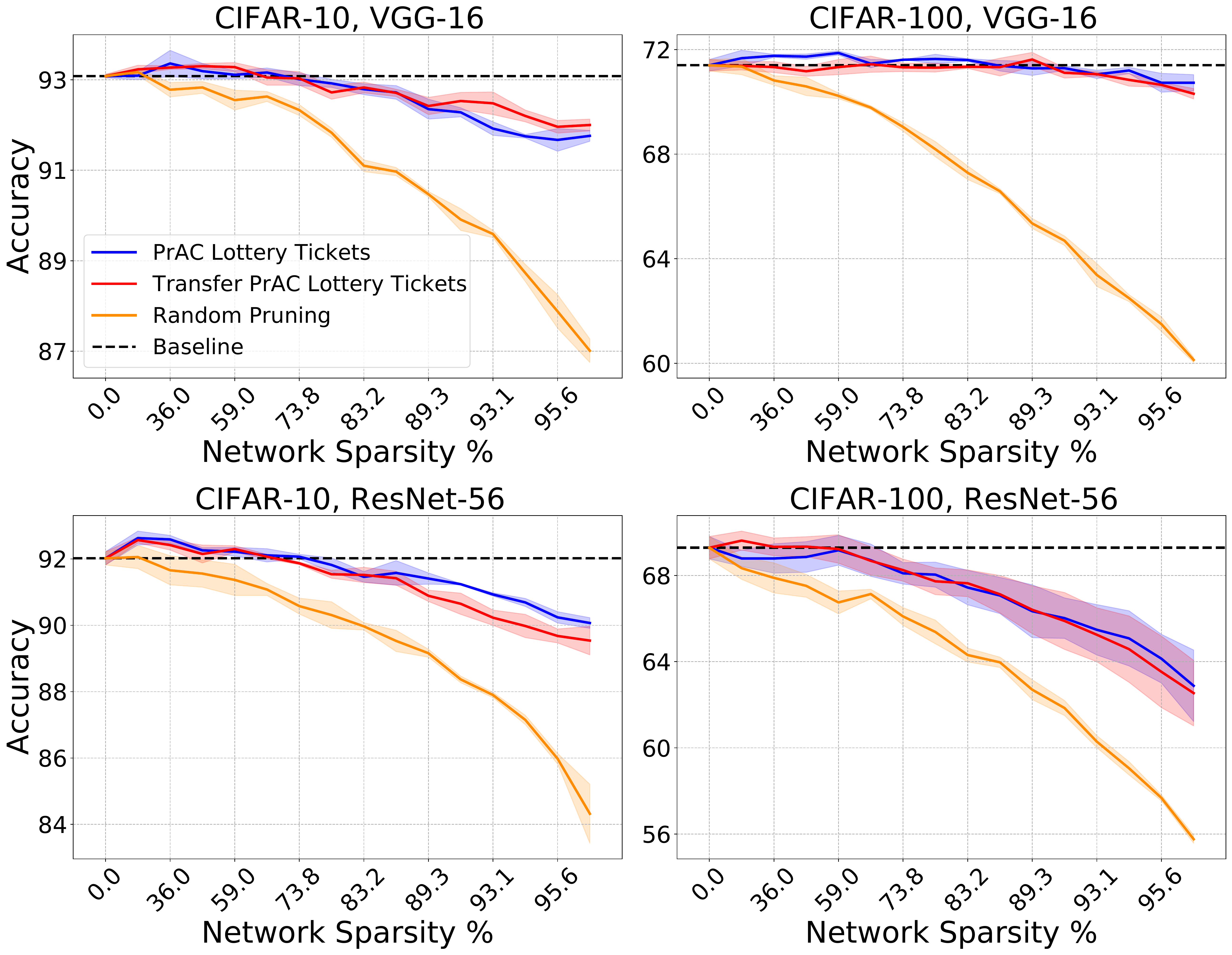}
 \vspace{-8mm} 
    \caption{The transferability study of PrAC sets on CIFAR-10 and CIFAR-100. \textcolor{blue}{Blue}, \textcolor{red}{Red},  \textcolor{orange}{Orange} and \textbf{Black} curves represent our PrAC lottery tickets, PrAC tickets found with transferred PrAC sets, random pruning and full network. Each curve contains the mean and standard deviation of test accuracy of subnetworks.}
    \vspace{-0.5em} 
    \label{fig:transfer}
\end{figure}

\vspace{-0.5em} 
\subsection{PrAC Sets Are Transferable Across Models}
\vspace{-0.5em} 
The construction of PrAC sets seems model-dependent, relying on a given full dense network and pruned subnetworks. It motives us to investigate to what extent the PrAC sets depend on those factors. As shown in Figure~\ref{fig:transfer}, we conduct transferability studies of PrAC sets across network architectures. Specifically, taking ResNet-20 as the source architecture to build PrAC sets on CIFAR-10 and CIFAR-100, and then finding PrAC lottery tickets in ResNet-56 and VGG-16 (target architectures) with transferred PrAC sets. 

% As mentioned in Section~\ref{sec:pac}, Pruning-Aware Critical set (PAC set) consists of both the hard samples that sparse network frequently forgets during training and the informative ones that contains the knowledge lost due to pruning. Although samples in the PAC set can help guide the model training, the process of constructing PAC set seems model-dependent. However, if the PAC set can transfer across different architectures, the resource consumption for constructing this set can be saved. Given the practical need, we conduct experiments to study the transferability of the PAC set between architectures. We choose ResNet-20 as the source architecture, construct PAC sets on CIFAR-10 and CIFAR-100, and use the identified PAC sets for ResNet-56 and VGG-16. 

Results in Figure~\ref{fig:transfer} demonstrate that \textit{Transfer PrAC Lottery Tickets} present competitive performance to \textit{PrAC Lottery Tickets}. They show similar accuracies at most sparsity levels, and both surpass randomly pruned subnetworks by a significant performance margin. It demonstrates that PrAC sets are surprisingly transferable for identifying lottery tickets across diverse architectures, which opens up promising avenues of efficiently finding winning tickets in huge models with compact PrAC sets constructed by tiny networks.

% using the PAC sets constructed on the source architecture, our method also achieve mathching performance as vanilla lottery tickets method. This suggests that our PAC set has great transferability that can share between different architectures and save unnecessary re-finding process. 
\vspace{-0.5em} 
\subsection{Comparisons with Strong Baselines.}

\begin{figure*}[!ht]
\vspace{-1.0em} 
    \centering
    \includegraphics[width=1.0\linewidth]{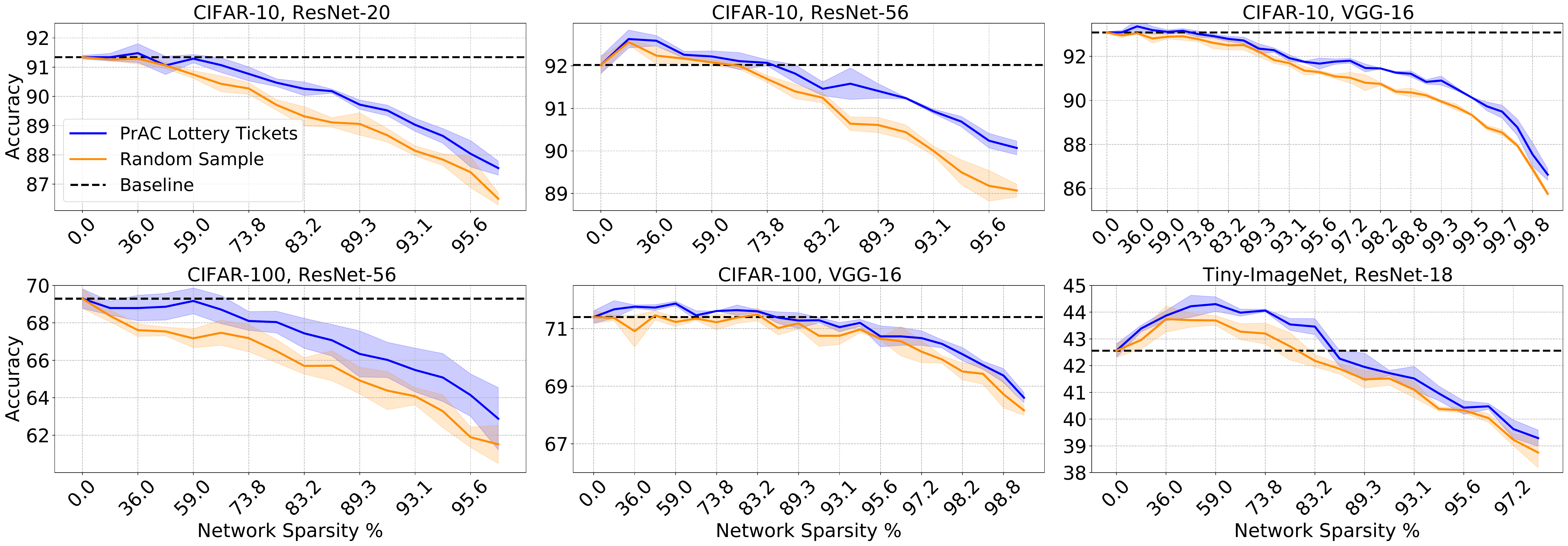}
    \vspace{-8mm} 
    \caption{Comparison results of our PrAC lottery tickets with subnetworks identified with subsets of random sampling across different architectures and datasets. \textbf{More results} can be found at Figure A10.}
    \vspace{-0.8em} 
    \label{fig:rand_sample}
\end{figure*}

\vspace{-0.5em} 
\paragraph{Core-set and active learning.}
% In previous paragraphs, we show that the PAC sets has contained most of the information for identifying winning tickets and have satisfactory transferability across different architectures. 
Natural comparative baselines, \textit{i.e.}, core-set and active learning approaches, are considered to assess the quality of PrAC sets further. In specific, we adopt two representative methods, active learning via maximum entropy sampling~\cite{lewis1994sequential,settles2012active} termed as ``Entropy", and core-set selection via proxy~\citep{Coleman2020Selection} named as ``SVP". Meanwhile, random sampling is designed for a sanity check.
% Maximum entropy-based sampling ranks the examples according to the entropy from a well trained network. Selection via proxy use a proxy network, which is obtained with much less training resources, to sample the examples with large entropy during training.\textcolor{Tianlong_color}{whether to detail the selection via proxy method, in the original paper they study 3 different methods, include forgetting events and max entropy.}
For fair comparisons, we keep the training iterations and the number of data in baselines consistent with our sparsity co-design approach. Figure~\ref{fig:ablation_datasample} collects the achieved performance of independently trained subnetworks from different approaches on CIFAR-10 with ResNet-20 and Figure~\ref{fig:rand_sample} further provides a comprehensive comparison with random sampling across different datasets and architectures. Results demonstrate that utilizing PrAC sets is capable of finding consistently better subnetworks with higher accuracies across diverse sparsity. It suggests that our co-design of data and model sparsity produces more informative pruning-aware subsets, which benefits to locate high-quality winning tickets. 

\begin{figure}[!ht]
    \centering
    \includegraphics[width=1.0\linewidth]{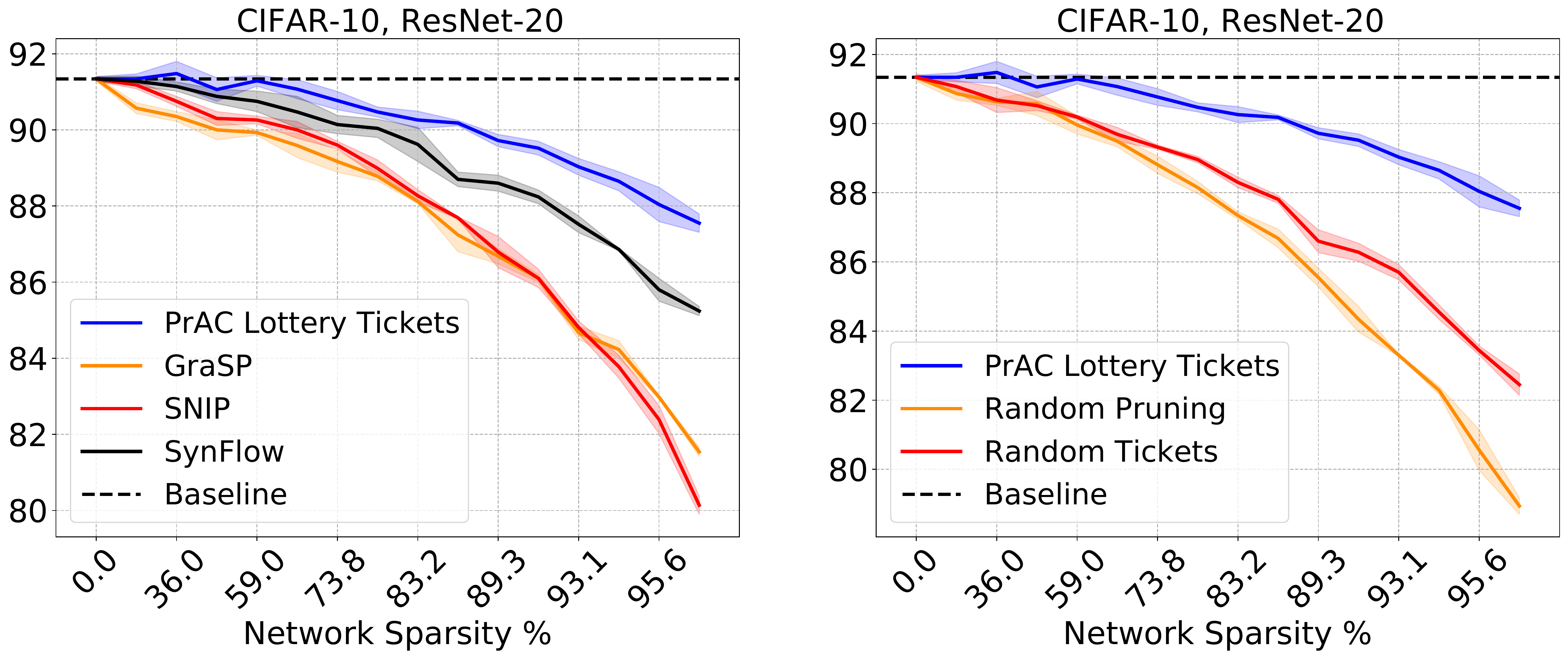}
    \vspace{-9mm} 
    \caption{Comparison results with strong baselines. \textit{Left:} Comparison of our PrAC lottery tickets with other pruning methods. \textit{Right:} Comparison of our methods with random pruning or initialization. }
    \label{fig:ablation_2}
    \vspace{-0.5em} 
\end{figure}

% Each curve contains the mean and standard deviation of testing accuracy of subnetworks at different sparsity levels.

\vspace{-0.5em}
\paragraph{Other efficient network pruning approaches.} Recent proposed SNIP~\citep{lee2018snip}, GraSP~\citep{Wang2020Picking}, and SynFlow~\citep{tanaka2020pruning} aim to prune networks at initialization, thereby saving resources at training stages. They usually only require a single batch of training examples with certain effective pruning criterion to find subnetworks in one-shot, which can be enhanced with more data and training budgets \cite{Wang2020Picking,tanaka2020pruning}. For fair comparisons, we implement these methods in an iterative manner (usually better than one-shot \cite{han2015deep,frankle2018lottery}) with the training iterations and the number of training data consistent with our approaches. As shown in Figure~\ref{fig:ablation_2} (\textit{Left}), only our approach is able to identify winning tickets with matched performance to full unpruned models (\textit{i.e.,} Baseline), and obtain a consistent performance margin compared to other pruning methods. Specifically, PrAC lottery tickets with $93.13\%$ sparsity surpass SynFlow, SNIP, and GraSP by $1.51\%$, $4.22\%$ and $4.36\%$ test accuracy. With the computation consumption remains constant for all algorithms, this achieved significant performance gap verifies the superiority of our proposal. 
% To further verify the training efficiency of our approach, we compare our algorithm with three state-of-the-art single-shot network pruning methods: SNIP~\citep{lee2018snip}, GraSP~\citep{wang2020picking} and SynFlow~\citep{tanaka2020pruning}. SNIP prunes the network based on the gradients of weights with respect to training loss. GraSP is designed to preserve gradient flow and uses Hessian-gradient product to score the weights. SynFlow aims to preserve the total flow of synaptic strengths in the network. The original setting of these three methods only require a single batch of images and have better performance when more data are available. 
\vspace{-1.3em} 
\paragraph{Random tickets with random re-initialization.} To exclude the possibility of trivial solutions, we consider the commonly adopted baseline, random tickets trained from randomly re-initialized weights, from the LTH literature \cite{frankle2018lottery}. From Figure~\ref{fig:ablation_2} (\textit{Right}), we observe that PrAC lottery tickets hold overwhelming advantages. For example, with a $1\%$ accuracy gap against the full model, our identified matching subnetworks with a sparsity of $79.03\%$, which are much sparser than both random pruning ($48.80\%$) and random tickets ($48.80\%$). 
% \cite{frankle2018lottery} claimed that winning tickets is a combination of weights and connections, which means both the position of pruned weights and the specific initialization are crucial for winning tickets. In this paragraph, we verify this statement in our data and model co-slimming framework. We compare our approach with two baseline methods: random tickets and random pruning~\citep{chen2020lottery}. ``Random tickets'' means that we reset the weights of the model to a new random initialization, and ``random pruning'' means that we randomly prune the model to the desired sparsity. 

% Moreover, our model can 
% we claim our methods find winning tickets as good as baseline lottery tickets with much less training resources.

\begin{figure}[!ht]
    \centering
    \includegraphics[width=1\linewidth]{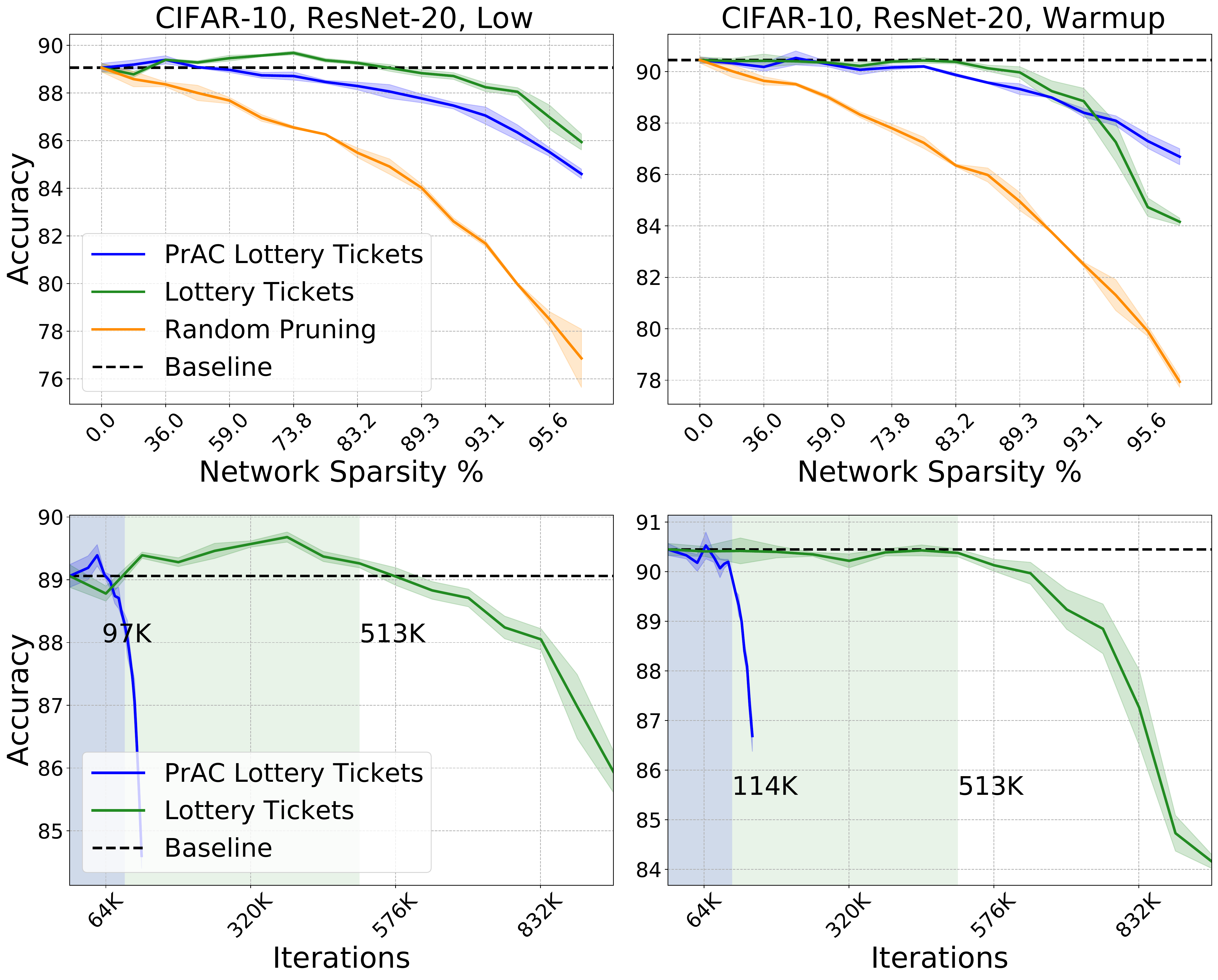}
    \vspace{-8mm}
    \caption{Testing accuracy of subnetworks at a range of sparsity levels from $0\%$ to $96.48\%$ (the first row) and the training iterations for finding each subnetwork (the second row) on CIFAR-10 with ResNet-20 under different lottery ticket settings.  The numbers within figures are the iterations used to find the subnetworks with the \textbf{same sparsity} and \textbf{comparable performance}. \textbf{More results} can be found at Figure A11.}
    \vspace{-4mm}
    \label{fig:ablation_low_wp}
\end{figure}
% \textcolor{blue}{Blue}, \textcolor{green}{Green}, \textcolor{orange}{Orange}, and \textbf{Black} curves represent our PrAC lottery tickets, vanilla lottery tickets, random pruning and full network, respectively.
% \vspace{-0.5em} 
\subsection{Ablation Study}
% \vspace{-0.5em} 
\paragraph{The Two Components in the PrAC set.} To investigate the individual effect of critical examples for training (CET) and pruning (CEP), we only utilize CET to identify matching subnetworks, as presented in Figure~\ref{fig:ablation_pac}. Results show that without the assistance of CEP, the found subnetworks consistently incur $\sim1\%$ performance drop. Table~\ref{tab:pac_number} collects the number of samples in CET and CEP. We observe that as the sparsity grows, the number of CEP keeps increasing; meanwhile, CEP shares fewer overlap images with CET, which indicates gradually detached distributions of critical samples during training and pruning. 

\begin{figure}[!ht]
\vspace{-0.5em} 
    \centering
    \includegraphics[width=1\linewidth]{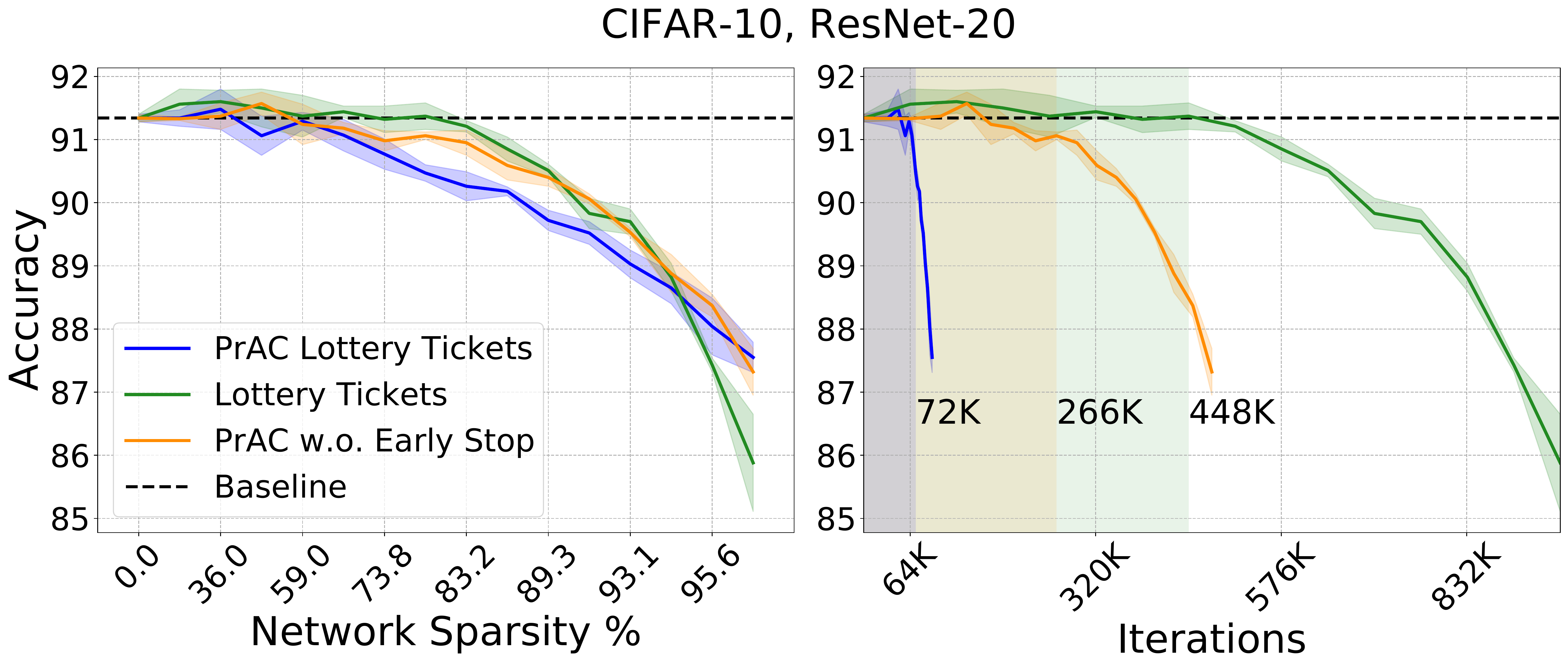}
    \vspace{-8mm} 
    \caption{Ablation studies of vanilla lottery tickets and our PrAC lottery tickets \textit{w/w.o.} early stopping on CIFAR-10 with ResNet-20. \textit{Left:} Testing accuracy of subnetworks with different sparsity. \textit{Right:} Training iterations for finding each subnetwork. And the numbers within the figure are the iterations used to find the corresponding subnetworks with the \textbf{same sparsity} and \textbf{comparable performance}. ($72$k, $266$k, $448$k represent PrAC lottery tickets, PrAC \textit{w.o.} Early Stop and vanilla lottery tickets, respectively.)}
    \label{fig:ablation_full}
    \vspace{-4mm} 
\end{figure}

\vspace{-1em} 
\paragraph{With or without early stopping.} We adopt the early stopping~\citep{You:2019tz} technique in our framework to find PrAC lottery tickets more efficiently. To understand its effect, we implement the variant, PrAC \textit{w.o.} Early Stop, that disables the early stopping in our methods. As shown in Figure~\ref{fig:ablation_full}, we observe that PrAC \textit{w.o.} Early Stop finds subnetworks with the same sparsity level and similar performance as vanilla lottery tickets, achieving $40.63\%$ training resources saving. Equipped with the early stopping, PrAC lottery tickets at the same sparsity, obtain $83.93\%$ training resources saving at the cost of $\le 0.50\%$ accuracy loss.

% reducing training effort has only slight negative effect on the performance of the identified matching subnetworks when using PAC sets (Specifically, $0.59\%$ lower in terms of testing accuracy at a sparsity of $79.03\%$). In general, all three subnetworks can reach a similar quality,
% %algorithms (PAC lottery ticket, find winning tickets of similar quality
% while two subnetworks found by our methods, PAC lottery ticket and PAC full training, can save $83.93\%$ and $40.63\%$ training resources of vanilla lottery tickets.

% \vspace{-0.5em} 
\paragraph{Validating across diverse lottery ticket settings.} Here we further evaluate our framework under two additional lottery ticket settings proposed by \citet{frankle2018lottery}, \textit{i.e.} \textbf{low} and \textbf{warmup}, with ResNet-20 and ResNet-56 on CIFAR-10 and CIFAR-100, respectively. Detailed hyperparameters are listed in Table~\ref{tab:setting}. Figure~\ref{fig:ablation_low_wp} and~\ref{fig:ablation_low_wp_100} shows that to find subnetworks with similar performance under the \textbf{low} and \textbf{warmup} settings, our methods only cost $18.91\%\sim22.22\%$ and $34.33\%\sim38.01\%$ training resources on CIFAR-10 and CIFAR-100, compared to vanilla lottery tickets. These consistently achieved training savings further verify the efficiency of PrAC lottery tickets, and the effectiveness of our sparsity co-design framework. 

% with a similar achieved performance of identified tickets, 
% while the training cost of our approach is only $18.91\%$ and $22.22\%$ of vanilla lottery tickets for \textbf{low} and \textbf{warmup} on CIFAR-10, and $34.33\%$ and $38.01\%$ on CIFAR-100. 

\begin{figure}[!ht]
% \vspace{-0.5em} 
    \centering
    \includegraphics[width=1\linewidth]{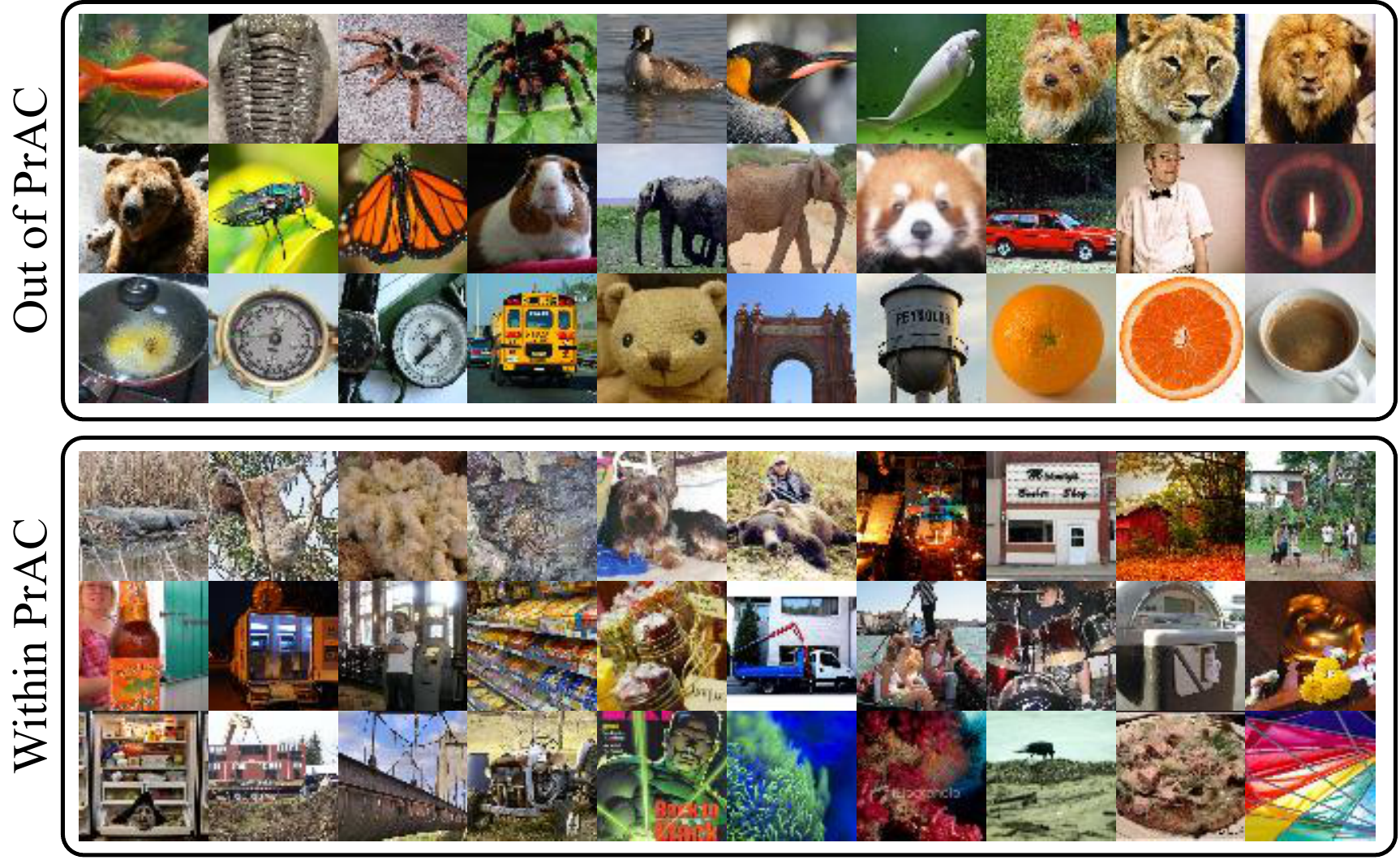}
    \vspace{-6mm} 
    \caption{Visualization of examples out of (\textit{upper}) and within (\textit{bottom}) the final PrAC set on Tiny-ImageNet.}
    \vspace{-4mm}
    \label{fig:pac}
%   \vspace{-0.5em} 
\end{figure}

% \vspace{-0.3em} 
\subsection{Visualization and Analyses of PrAC Sets}
% \vspace{-0.2em} 
Visualization of examples out of and within PrAC sets on Tiny-ImageNet is provided in Figure~\ref{fig:pac}, and the class-wise ratios of images in PrAC sets can be found in Figure~\ref{fig:class_wise_pac}. As shown in Figure~\ref{fig:pac}, the images out of the PrAC set show less complexity where objects are centered and easily distinguishable, such as identifying an orange from white backgrounds. In contrast, the images within PrAC sets contain multiple ambiguous elements, including lower quality, depict multiple objects, complicated backgrounds and resulting in a challenging recognition even for a human. In addition, the distribution of PrAC set's classes in Figure~\ref{fig:class_wise_pac} is quite balanced, where the number of images is in the same order. Such observations may provide possible insights on why PrAC sets are capable of locating critical subnetworks, \textit{i.e.}, PrAC tickets, with satisfying performance.

% \vspace{-0.1em} 
\section{Conclusion}
% \vspace{-0.5em} 
In this paper, we explore a new perspective to finding lottery tickets more efficiently by doing so on small pruning-aware critical (PrAC) subsets, which is constructed via data and model sparsity co-design. Extensive experiments verify the effectiveness of our proposals with diverse network architectures on multiple common datasets, including CIFAR-10, CIFAR-100, and Tiny ImageNet. High-quality winning tickets, can be identified efficiently on such compact PrAC sets and enjoys significant training cost reduction.

\clearpage

\bibliography{DMCLTH}
\bibliographystyle{icml2021}

\clearpage

\appendix
\renewcommand{\thepage}{A\arabic{page}}  
\renewcommand{\thesection}{A\arabic{section}}   
\renewcommand{\thetable}{A\arabic{table}}   
\renewcommand{\thefigure}{A\arabic{figure}}

\section{More Implementation Details} \label{appendix_setting} 
\paragraph{Training and evaluation details.} We use an SGD optimizer with a momentum of $0.9$ and a weight decay of $10^{-4}$ in our experiments. And we choose the model with the best validation accuracy during the training process. Besides, we use an early weight rewinding~\citep{frankle2019linear} method (rewind to the third epoch) to help scale up the lottery ticket hypothesis in these models, except for the warmup and low variant of ResNet-20 and ResNet-56, in which the weight will be rewound to the same random initialization. For the variant of warmup, we replace the original $85$ epochs~\citep{frankle2018lottery} with $15$ epochs, which does not affect the performance. The threshold of the number of forgets is set to $0$ and we default to use $0.07$ as the threshold for the distance between masks. Note that our baseline results are aligned with \cite{frankle2018lottery}.

\paragraph{Dataset.} We consider three datasets in our implementation, which can be download at \url{https://www.cs.toronto.edu/~kriz/cifar.html} for CIFAR-10 and CIFAR-100, and \url{http://cs231n.stanford.edu/tiny-imagenet-200.zip} for Tiny-ImageNet. For all three datasets, $10$ percent of data from the training set are randomly split up as validation set. And we utilize random cropping and random horizontal flipping for data augmentation.
\paragraph{Computing infrastructures.}
All our experiments are conducted on Quadro RTX 6000 and Tesla V100 GPUs.

\section{More Experiment Results}
\subsection{More Results of Sampling Strategy}
\begin{figure}[!ht]
    \centering
    \includegraphics[width=1\linewidth]{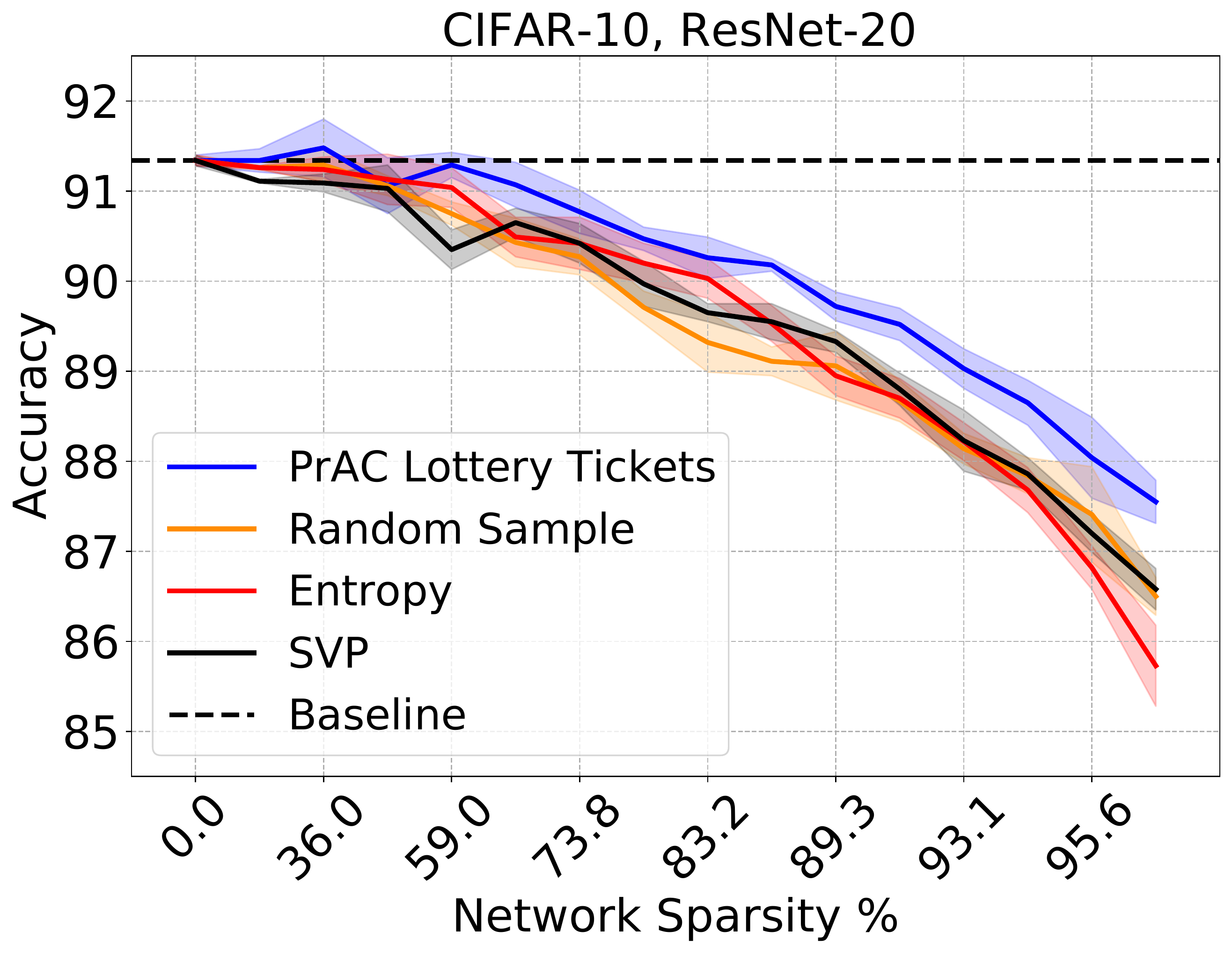}
    \vspace{-6mm}
    \caption{Comparison of our PrAC sets with other core-sets or active learning approaches.}
    \label{fig:ablation_datasample}
\end{figure}

As shown in Figure~\ref{fig:ablation_datasample}, our PrAC sets achieve consistent improvement compare with other sampling strategies. It indicates that our approach produces more informative pruning-aware subsets and contribute for finding high-quality winning tickets.

\subsection{More Results of Different Lottery Ticket Settings}
\begin{figure}[!ht]
    \centering
    \includegraphics[width=1\linewidth]{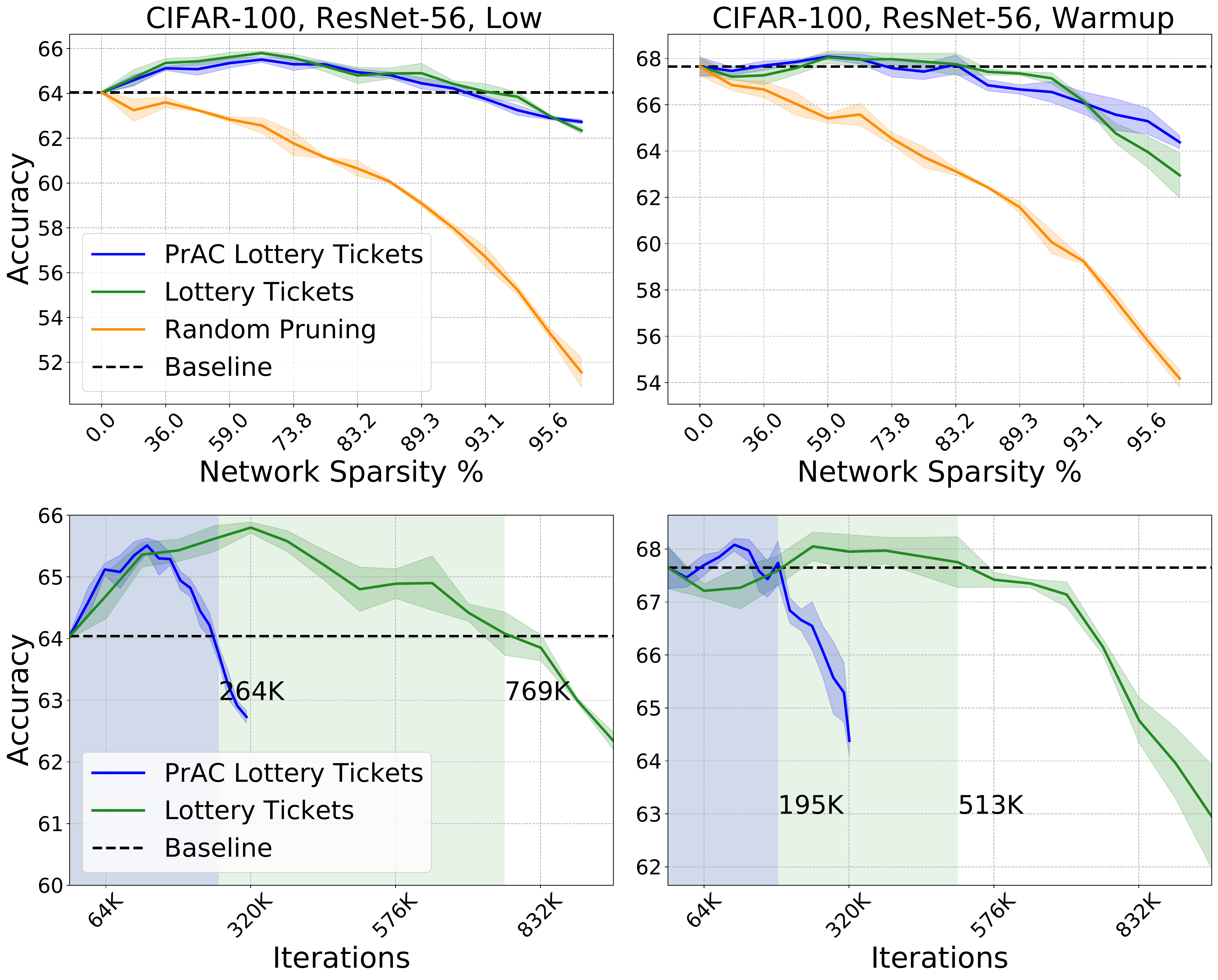}
    \vspace{-8mm}
    \caption{Testing accuracy of subnetworks at a range of sparsity levels from $0\%$ to $96.48\%$ (the first row) and the training iterations for finding each subnetwork (the second row) on CIFAR-100 with ResNet-56 under different lottery ticket settings. \textcolor{blue}{Blue}, \textcolor{green}{Green}, \textcolor{orange}{Orange}, and \textbf{Black} curves represent our PrAC lottery tickets, vanilla lottery tickets, random pruning and full network, respectively. The numbers within figures are the iterations used to find the subnetworks with the \textbf{same sparsity} and \textbf{comparable performance}.}
    \label{fig:ablation_low_wp_100}
\end{figure}

Figure~\ref{fig:ablation_low_wp_100} reports the performance on CIFAR-100 with ResNet-56 under two additinal lottery tickets settings, \textbf{low} and \textbf{warmup}. We can observe that our methods cost $34.33\%\sim38.01\%$ training sources and achieve comparable performance, which suggests the efficiency of our PrAC lottery tickets.

\subsection{More Statistics of PrAC Sets}
Table~\ref{tab:pac_size} contains the size of PrAC sets across different datasets and networks. On CIFAR-10 ($10$ classes), we locate PrAC sets with the size range from $35.32\%$ to $37.07\%$ of the training set, while $69.55\%$ to $78.19\%$ on CIFAR-100 ($100$ classes) and $68.23\%$ to $75.10\%$ on Tiny-ImageNet ($200$ classes). The result suggests that more data are needed to find high-quality PrAC lottery tickets for the image recognition with more classes.
\begin{table}[!ht]
    \centering
    \vspace{-2mm}
    \caption{Proportion of PrAC sets to their training set sizes of CIFAR-10, CIFAR-100 and Tiny-ImageNet}
    \label{tab:pac_size}
    \resizebox{0.48\textwidth}{!}{
    \begin{tabular}{c|c|c}
        \toprule
         Dataset & Network & Proportion of PrAC sets  \\ \midrule
          & ResNet-20 & $36.66\%$  \\
         CIFAR-10 & ResNet-56 & $37.07\%$ \\
          & VGG-16 & $ 35.32\%$\\ \midrule
          
          & ResNet-20 & $78.19\%$  \\
         CIFAR-100 & ResNet-56 & $74.94\%$ \\
          & VGG-16 & $ 69.55\%$\\ \midrule          
         
         \multirow{2}*{Tiny-ImageNet}& ResNet-18 & $75.10\%$ \\
         ~ & VGG-16 & $68.23\%$  \\
 \bottomrule
    \end{tabular}}
    \vspace{-2mm}
\end{table}

\subsection{More Results of Ablation Study}

\begin{figure}[!ht]
    \centering
    \includegraphics[width=1\linewidth]{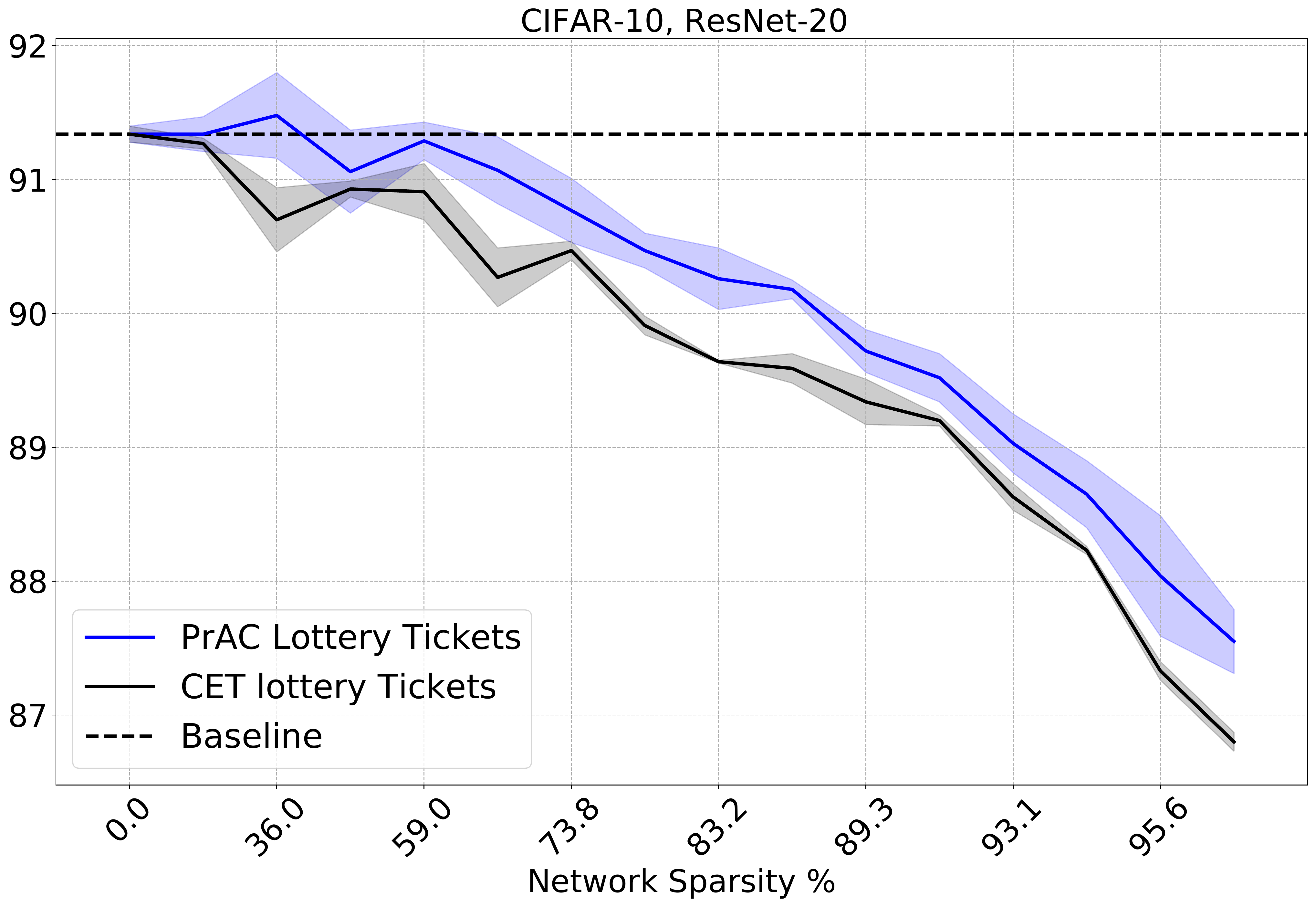}
    \vspace{-6mm}
    \caption{Comparison of PrAC lottery tickets with CET lottery tickets on CIFAR-10 with ResNet-20. Each curve contains the mean and standard deviation of testing accuracy of subnetworks at different sparsity levels.}
    \label{fig:ablation_pac}
\end{figure}

\paragraph{The two components in the PrAC set.} We conduct our data and model co-design framework with only critical examples for training (CET), named as CET lottery tickets. As shown in Figure~\ref{fig:ablation_pac}, without the assistance of critical examples for pruning (CEP), there is a consistent performance gap between PrAC lottery tickets and CET lottery tickets. Besides, we collect the number of CET, CEP and PrAC sets in Table~\ref{tab:pac_number}. The overlapping rate means the percentage of the overlap images between CET and CEP sets in CEP sets, (\textit{i.e.}, $\frac{|\mathrm{CEP}| \cap |\mathrm{CET}|}{|\mathrm{CEP}|}$). We observe that as the sparsity grows, the number of CEP sets increases while the overlapping rate decreases, which indicates gradually detached distributions of critical samples during training and pruning.

\begin{table}[!ht]
    \centering
    \vspace{-2mm}
    \caption{Results of the number of the identified CET, CEP and PrAC sets, as well as the overlapping rate of CEP sets during the process of our co-design framework on CIFAR-10 with ResNet-20.}
    \label{tab:pac_number}
    \resizebox{0.48\textwidth}{!}{
    \begin{tabular}{c|c|c|c|c}
        \toprule
        Sparsity of Subnetworks & CEP & CET & PrAC & Overlapping Rate  \\ \midrule
        $20.00\%$ & $1501$ & $24159$ & $24168$ & $99.40\%$ \\
        $36.00\%$ & $1481$ & $21708$ & $21728$ & $98.65\%$ \\
        $48.80\%$ & $3935$ & $19542$ & $19838$ & $92.48\%$ \\
        $59.04\%$ & $3782$ & $17674$ & $18161$ & $87.12\%$ \\
        $67.23\%$ & $4723$ & $16091$ & $16712$ & $86.85\%$ \\
        $73.79\%$ & $5514$ & $14771$ & $16026$ & $77.24\%$ \\
        $79.03\%$ & $4420$ & $14357$ & $15202$ & $80.88\%$ \\
        $83.22\%$ & $4602$ & $13909$ & $14880$ & $78.90\%$ \\
        $86.58\%$ & $5391$ & $13741$ & $14980$ & $77.02\%$ \\
        $89.26\%$ & $5360$ & $14168$ & $15376$ & $77.46\%$ \\
        $91.41\%$ & $5098$ & $14365$ & $15247$ & $82.70\%$ \\
        $93.13\%$ & $5840$ & $14553$ & $15804$ & $78.58\%$ \\
        $94.50\%$ & $5728$ & $14959$ & $16062$ & $80.74\%$ \\
        $95.60\%$ & $6370$ & $15290$ & $16360$ & $83.20\%$ \\
        $96.48\%$ & $6616$ & $15369$ & $16499$ & $82.92\%$ \\
        \bottomrule
    \end{tabular}}
    \vspace{-2mm}
\end{table}

\begin{figure}[!ht]
    \centering
    \includegraphics[width=1\linewidth]{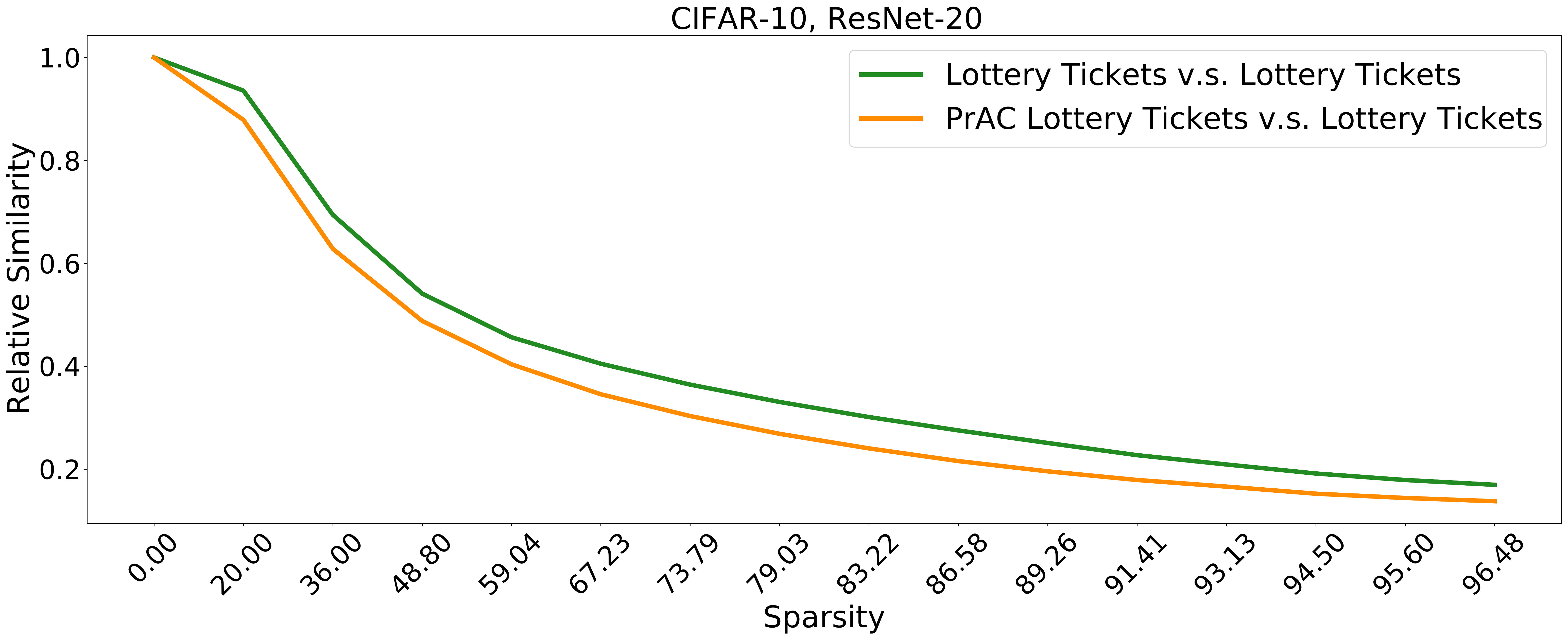}
    \vspace{-6mm}
    \caption{Results of the relative mask similarity on CIFAR-10 with ResNet-20. \textcolor{green}{Green} and \textcolor{orange}{Orange} represents the relative similarity between two independent runs of vanilla lottery tickets, and the one between PrAC lottery tickets and vanilla lottery tickets. We adopt the same random initialization for identifying these three groups of subnetworks.}
    \label{fig:relative_similarity}
    \vspace{-4mm}
\end{figure}

\paragraph{Relative similarity between PrAC LT and LT.} We evaluate the overlap degree in sparsity patterns with relative similarity (\textit{i.e.}, $\frac{m_i \cap m_j}{m_i \cup m_j}$), where $m_i$ and $m_j$ are the sparsity masks of identified subnetworks. We keep the same random initialization for PrAC lottery tickets and two independent runs of vanilla lottery tickets. Figure~\ref{fig:relative_similarity} shows that as the sparsity grows, subnetworks share fewer sparsity patterns. And the relative similarity between PrAC lottery tickets and lottery tickets are slightly smaller than between two different runs of lottery tickets, which indicates the non-trivial difference between sparse masks of PrAC LT and LT.

\begin{figure}[!ht]
    \centering
    \includegraphics[width=1\linewidth]{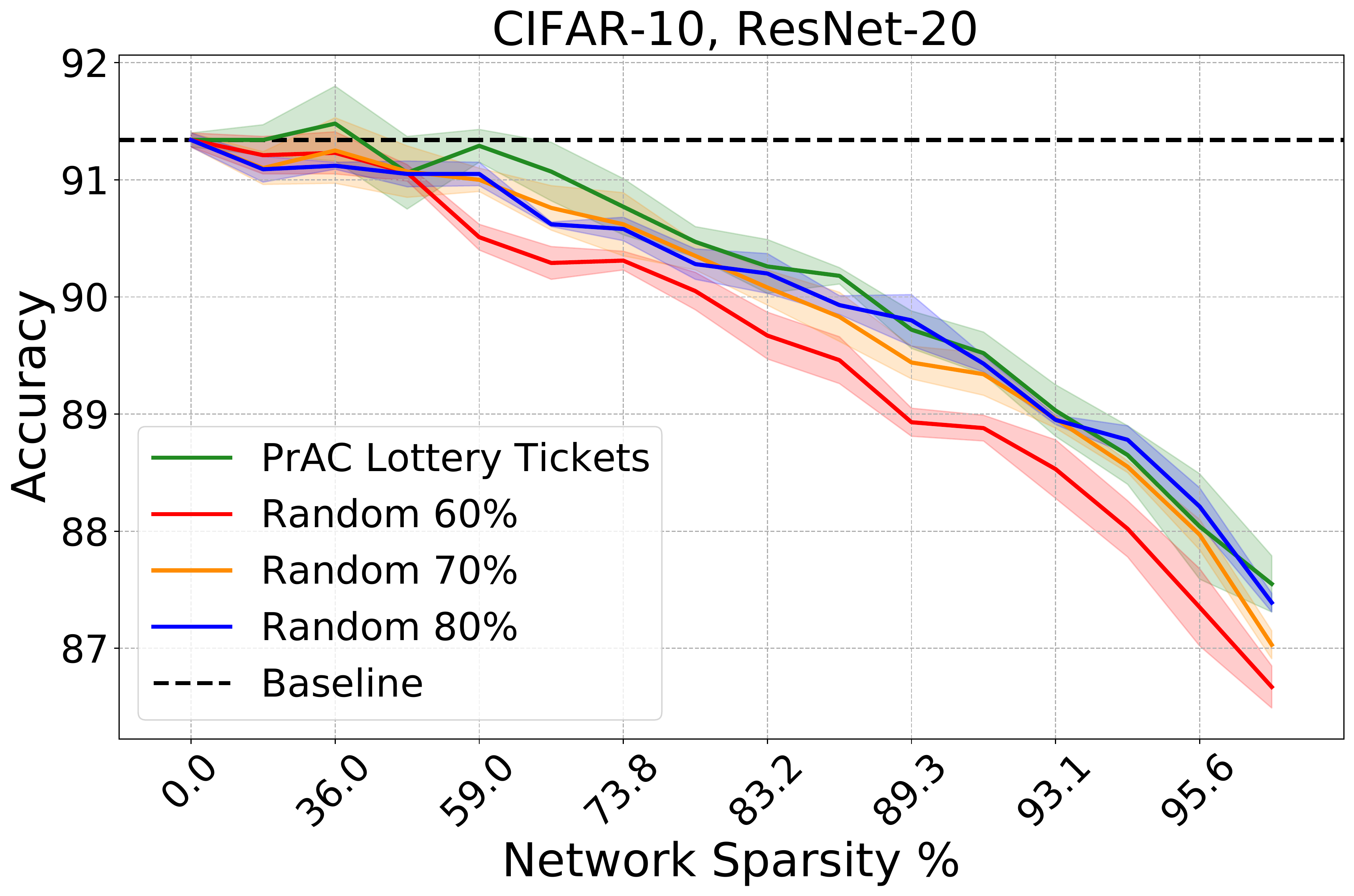}
    \vspace{-6mm}
    \caption{Comparison of the quality of subnetworks identified by our PrAC subsets and subsets from random sampling on CIFAR-10 with ResNet-20. We keep the size of random subsets consistent during the whole IMP process, ranging from $60\%\sim80\%$.}
    \label{fig:fix_rand}
    \vspace{-4mm}
\end{figure}

\paragraph{Lottery tickets with subsets of random sampling} To investigate that how many examples of random sampling can match the performance of our PrAC subsets in terms of locating subnetworks, we conduct an ablation study on CIFAR-10 with ResNet-20 and record the results in Figure~\ref{fig:fix_rand}. We can observe that nearly $70\%$ data are needed for random subsets to match the performance of our PrAC sets, which only contain $37\%\sim54\%$ data. 

\begin{table}[!ht]
    \centering
    \vspace{-2mm}
    \caption{Results of test accuracy of identified subnetworks with respect to the threshold for the number of forgets on CIFAR-10 with ResNet-20. We select subnetworks with the same sparsity of $16.78\%$, which is the maximum sparsity of subnetworks identified by PrAC subsets ($\mathcal{E}_{\mathrm{F}}=0$) have \textbf{comparable performance}.}
    \label{tab:thres}
    \resizebox{0.48\textwidth}{!}{
    \begin{tabular}{c|cccccc}
        \toprule
        $\mathcal{E}_{\mathrm{F}}$ & $0$ & $2$ & $4$ & $6$ & $8$ & $10$ \\ \midrule
        Accuracy ($\%$) & $91.05$ & $90.43$ & $90.35$ & $89.32$ & $89.08$ & $88.79$ \\ \midrule
        PrAC & $19748$ & $14992$ & $12536$ & $11141$ & $11152$ & $10338$ \\ \bottomrule
    \end{tabular}}
    \vspace{-2mm}
\end{table}

\paragraph{The threshold for the number of forgets} Table~\ref{tab:thres} records the test accuracy and the size of PrAC subsets under different threshold for the number of forgets. We can observe that both the test accuracy and the size of PrAC decrease as the threshold rises. Thus we choose $\mathcal{E}_{\mathrm{F}}=0$ in our implementation.

\subsection{More Visualization and Analyses}
\begin{figure}[!ht]
    \centering
    \includegraphics[width=1\linewidth]{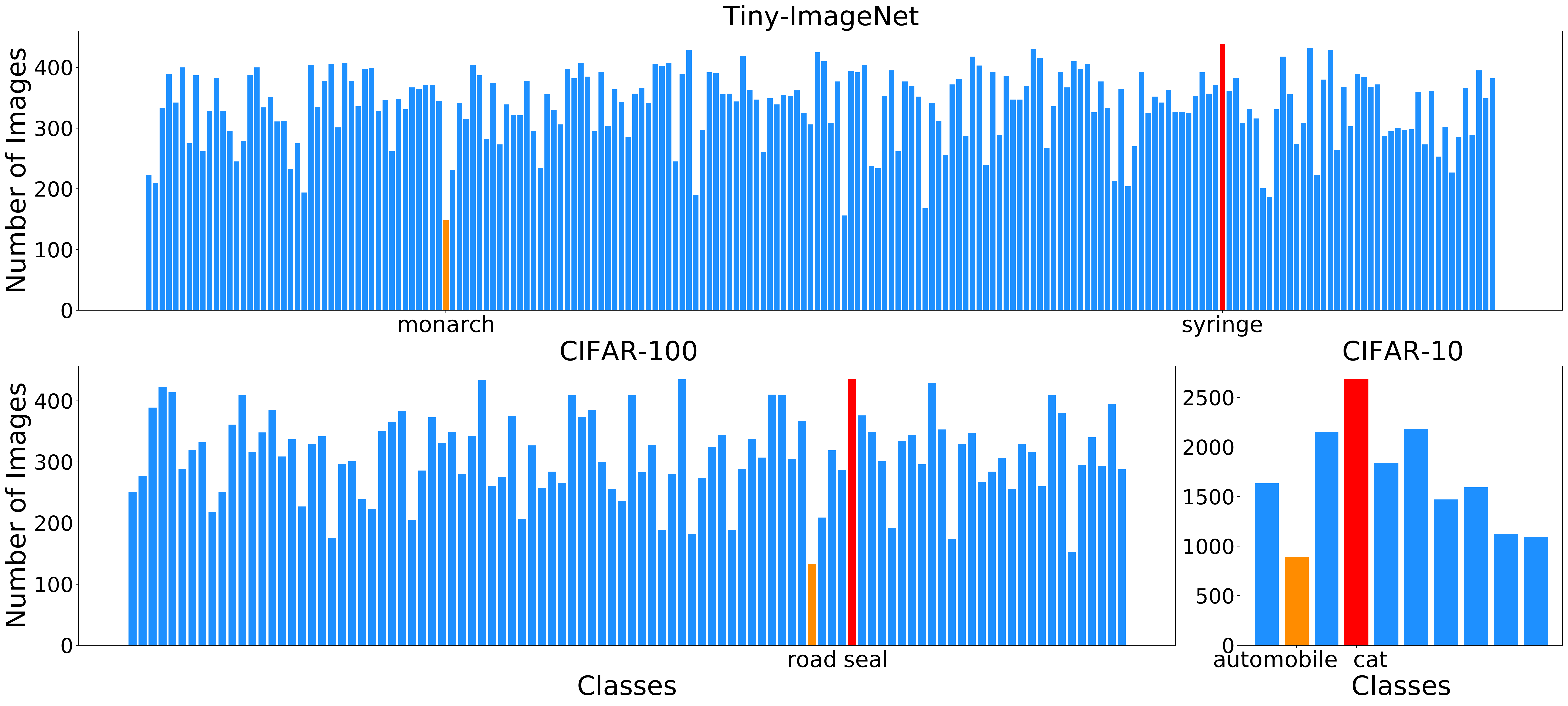}
    \vspace{-6mm}
    \caption{The class-wise ratios of images in PrAC sets on CIFAR-10/100 and Tiny-ImageNet, respectively. \textcolor{red}{Red} and \textcolor{orange}{Orange} represent the classes with maximum and minimum images.}
    \label{fig:class_wise_pac}
    \vspace{-2mm}
\end{figure}
Figure~\ref{fig:class_wise_pac} demonstrates the class-wise ratios of images in PrAC set, from which we can find that the number of images from different classes are in the same order. This balanced distribution of PrAC set's classes may provide possible insights on the effectiveness of PrAC sets, with respect to locating critical subnetworks, i.e., PrAC tickets, with satisfying performance.

\subsection{Additional Results of Forgetting Statistics in LT}

\begin{figure}[!h]
    \centering
    \includegraphics[width=0.77\linewidth]{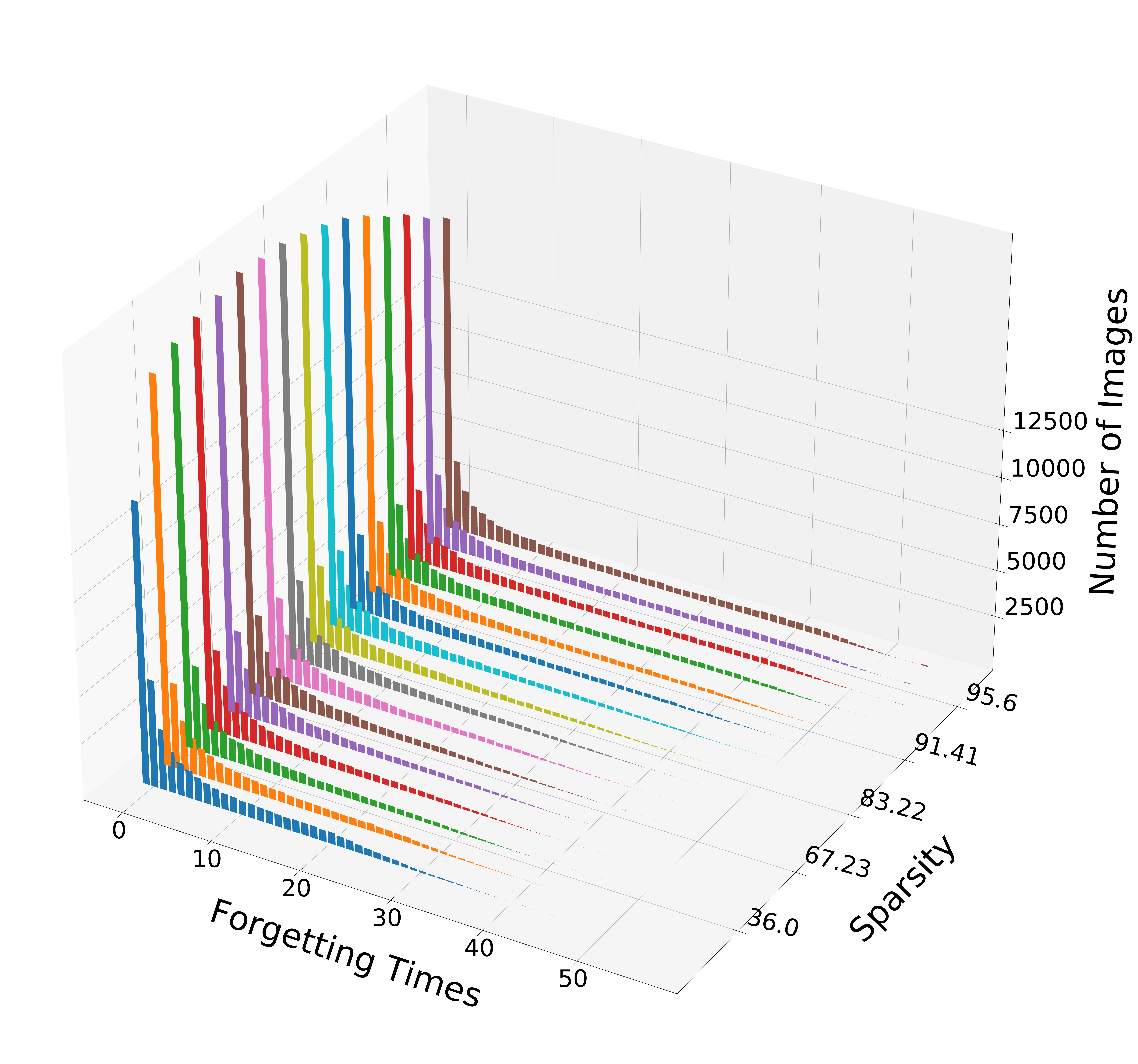}
    \includegraphics[width=0.77\linewidth]{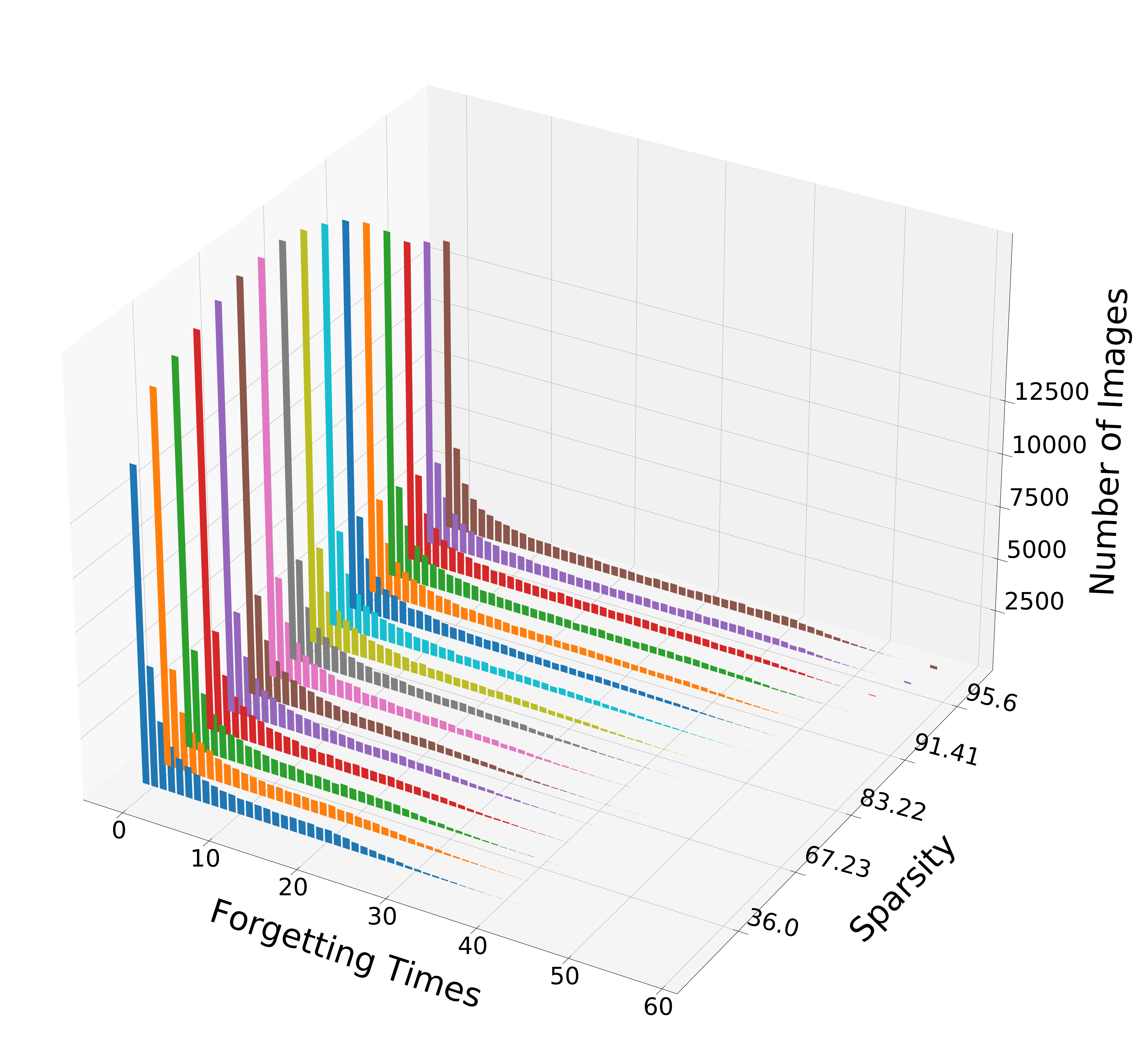}
    \includegraphics[width=0.77\linewidth]{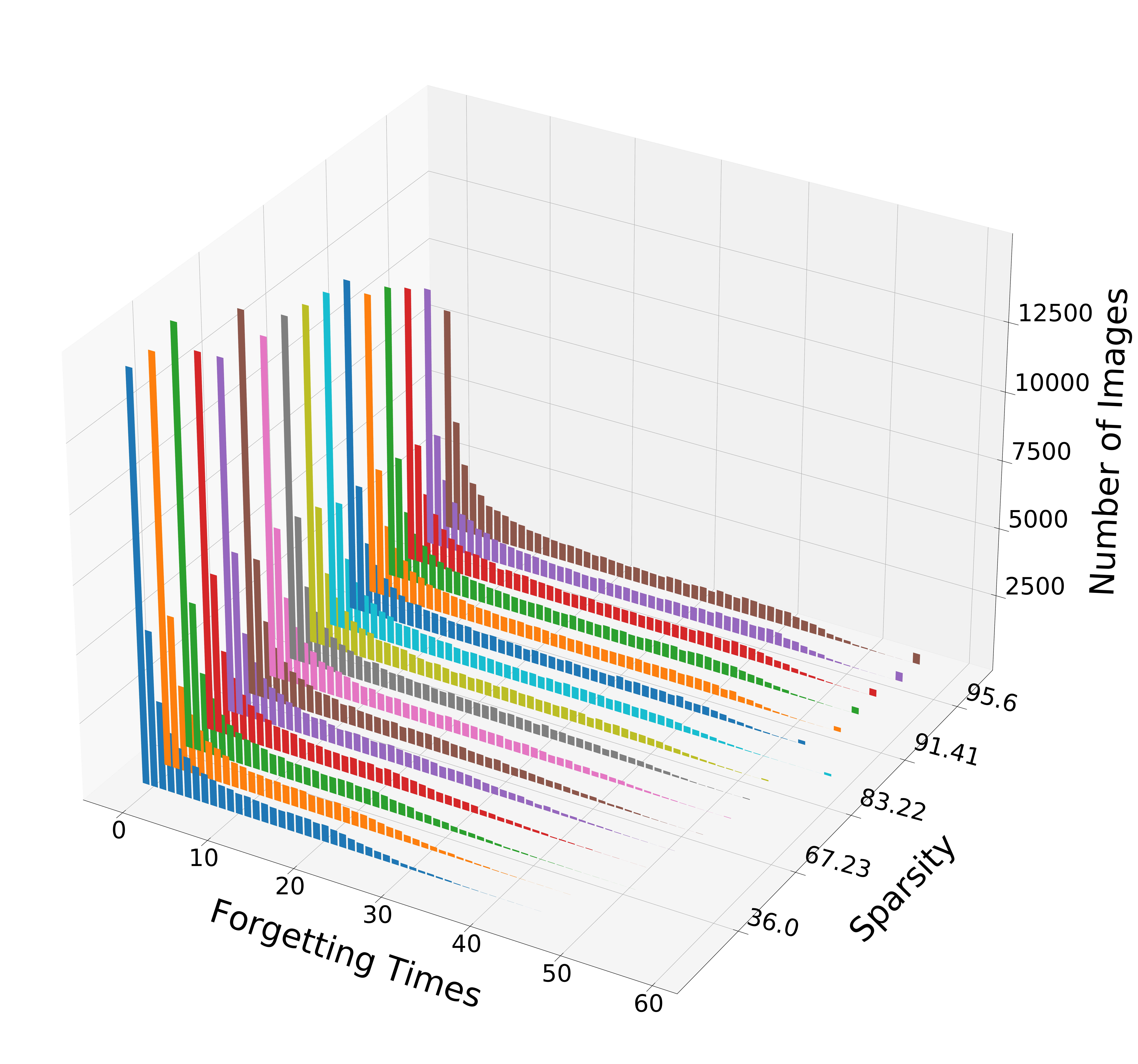}
    \vspace{-2mm}
    \caption{Visualization of the forgetting statistics of subnetworks at different sparsity from $0\%$ to $96.48\%$ on CIFAR-10 with ResNet-20 when training with full data. \textit{Top:} Basic iterative magnitude pruning (fine-tune after pruning). \textit{Middle:} vanilla lottery tickets. \textit{Bottom:} random tickets.}
    \label{fig:CET}
    \vspace{-3mm}
\end{figure}

Figure~\ref{fig:CET} shows the distribution of training data's forgetting times at different sparsity from $0\%$ to $96.48\%$ on CIFAR-10 with ResNet-20. We consider three pruning methods: Basic iterative magnitude pruning (IMP) \cite{han2015deep}, vanilla lottery tickets (LT) \cite{frankle2018lottery}, and random tickets (RT). IMP fine-tune the subnetworks directly after pruning while LT rewinds the weight to the same initialization and RT reinitializes the subnetworks before fine-tuning. We can observe that as the sparsity increases, for IMP and LT, the number of unforgettable images first increases and then decreases, while the one for RT consistently decreases. Besides, the maximum number of forgetting times grows as the sparsity becomes larger. 

\end{document}